\documentclass[journal,twoside]{IEEEtran}
\usepackage{graphicx}
\usepackage{multirow}
\usepackage{textcomp}
\usepackage{subcaption}
\usepackage{amsmath}
\usepackage{breqn}
\usepackage{mathtools}
\usepackage{lipsum}
\usepackage{pdflscape}
\usepackage{tabularx,booktabs, longtable}
\begin{document}

\title{Epileptic Seizures Detection Using Deep Learning Techniques: A Review}

\author{Afshin~Shoeibi,
        Marjane~Khodatars,
        Navid~Ghassemi,
        Mahboobeh~Jafari,
        Parisa~Moridian,
        Roohallah~Alizadehsani,
        Maryam~Panahiazar,
        Fahime~Khozeimeh,
        Assef~Zare,
        Hossein~Hosseini-Nejad,
        Abbas~Khosravi,
        Amir~F.~Atiya,
        Diba~Aminshahidi,
        Sadiq~Hussain,
        Modjtaba~Rouhani,
        Saeid~Nahavandi,
        and~Udyavara~Rajendra~Acharya% <-this % stops a space
\thanks{A. Shoeibi and N. Ghassemi are with the Faculty of Electrical Engineering, Biomedical Data Acquisition Lab (BDAL), K. N. Toosi University of Technology, Tehran 1631714191, Iran, and the Computer Engineering Department, Ferdowsi University of Mashhad, Mashhad 9177948974, Iran. (Corresponding author: Afshin Shoeibi, email: afshin.shoeibi@gmail.com).}% <-this % stops a space
\thanks{M. Khodatars is with the Mashhad Branch, Islamic Azad University, Mashhad 91735413, Iran.}
\thanks{M. Jafari is with Electrical and Computer Engineering Faculty, Semnan University, Semnan 3513119111, Iran.}
\thanks{P. Moridian is with the  Faculty of Engineering, Science and Research Branch, Islamic Azad University, Tehran 1477893855, Iran.}
\thanks{R. Alizadehsani, F. Khozeimeh, A. Khosravi and S. Nahavandi. are with the Institute for Intelligent Systems Research and Innovation (IISRI), Deakin University, Waurn Ponds, VIC 3217, Australia.}
\thanks{M. Panahiazar is with the Institute for Computational Health Sciences, School of Medicine, University of California, San Francisco, CA 94143, USA.}
\thanks{A. Zare is with  Faculty of Electrical Engineering, Gonabad Branch, Islamic Azad University, Gonabad 6518115743, Iran.}
\thanks{H. Hosseini-Nejad is with the Faculty of Electrical and Computer Engineering, K. N. Toosi University of Technology, Tehran 1631714191, Iran.}% <-this % stops a space
\thanks{A. F. Atiya is with the Department of Computer Engineering, Faculty of Engineering, Cairo University, Cairo 12613, Egypt.}
\thanks{D. Aminshahidi and M. Rouhani are with the Computer Engineering Department, Ferdowsi University of Mashhad, Mashhad 9177948974, Iran.}
\thanks{S. Hussain is System Administrator at Dibrugarh University, Assam 786004, India.}
\thanks{U. R. Acharya is with the  Department of Biomedical Engineering, School of Science and Technology, Singapore University of Social Sciences, Singapore 599494, Singapore, the Department of Electronics and Computer Engineering, Ngee Ann Polytechnic, Singapore 599489, Singapore, and the Department of Bioinformatics and Medical Engineering, Taichung City 41354, Taiwan.}% <-this % stops a space
% \thanks{Manuscript received April 19, 2005; revised August 26, 2015.}
}

%\markboth{IEEE Transactions on Neural Networks and Learning Systems}{Shoeibi \MakeLowercase{\textit{et al.}}: Application of deep learning techniques for automated detection of epileptic seizures}

\maketitle

\begin{abstract}
A variety of screening approaches have been proposed to diagnose epileptic seizures, using electroencephalography (EEG) and magnetic resonance imaging (MRI) modalities. Artificial intelligence encompasses a variety of areas, and one of its branches is deep learning (DL). Before the rise of DL, conventional machine learning algorithms involving feature extraction were performed. This limited their performance to the ability of those handcrafting the features. However, in DL, the extraction of features and classification are entirely automated. The advent of these techniques in many areas of medicine, such as in the diagnosis of epileptic seizures, has made significant advances. In this study, a comprehensive overview of works focused on automated epileptic seizure detection using DL techniques and neuroimaging modalities is presented. Various methods proposed to diagnose epileptic seizures automatically using EEG and MRI modalities are described. In addition, rehabilitation systems developed for epileptic seizures using DL have been analyzed, and a summary is provided. The rehabilitation tools include cloud computing techniques and hardware required for implementation of DL algorithms. The important challenges in accurate detection of automated epileptic seizures using DL with EEG and MRI modalities are discussed. The advantages and limitations in employing DL-based techniques for epileptic seizures diagnosis are presented. Finally, the most promising DL models proposed and possible future works on automated epileptic seizure detection are delineated.
\end{abstract}


\IEEEpeerreviewmaketitle

\section{Introduction}

\IEEEPARstart{E}{pilepsy} is a non-communicable disease and one of the most common neurological disorders of humans, usually associated with sudden attacks \cite{one}. Seizures are a swift and early abnormality in the electrical activity of the brain that disrupts the part of the whole body \cite{two}. Epileptic seizures are affecting around 60 million people worldwide by varied kinds \cite{three}. These attacks occasionally provoke cognitive disorders which can cause severe physical injury to the patient. Besides, people with epileptic seizures sometimes suffer emotional distress due to embarrassment and lack of appropriate social status. Hence, early detection of epileptic seizures can help the patients and improve their quality of life. 

Various screening techniques have been developed to diagnose epileptic seizures, including MRI \cite{four}, EEG \cite{five}, Magnetoencephalography (MEG) \cite{six} and Positron Emission Tomography (PET) \cite{seven}. The EEG signals are widely preferred as they are economical, portable, and show clear rhythms in the frequency domain \cite{eight}. The EEG provides the voltage variations produced by the ionic current of neurons in the brain, which indicate the brain's bioelectric activity \cite{nine}. Diagnosing epilepsy with EEG signals is time-consuming and strenuous, as the epileptologist or neurologist needs to screen the EEG signals minutely. Also, there is a possibility of human error, and hence, developing a computer-based diagnosis may alleviate these problems. 

Many machine learning algorithms have been developed using statistical, frequency domain and nonlinear parameters to detect epileptic seizures \cite{ten,eleven,twelve,thirteen,fourteen,fifteen}. In conventional machine learning techniques, the selection of features and classifiers is made by trial and error method. Also, one needs to have sound knowledge of signal processing and data mining to develop a robust model. Such models perform well for limited data. Nowadays, with the increase in the availability of data, machine learning techniques may not perform very well. Hence, the deep learning techniques, which are the state-of-art methods, have been employed. 

\begin{figure}[t]
    \centering
    \includegraphics[width=3.5in ]{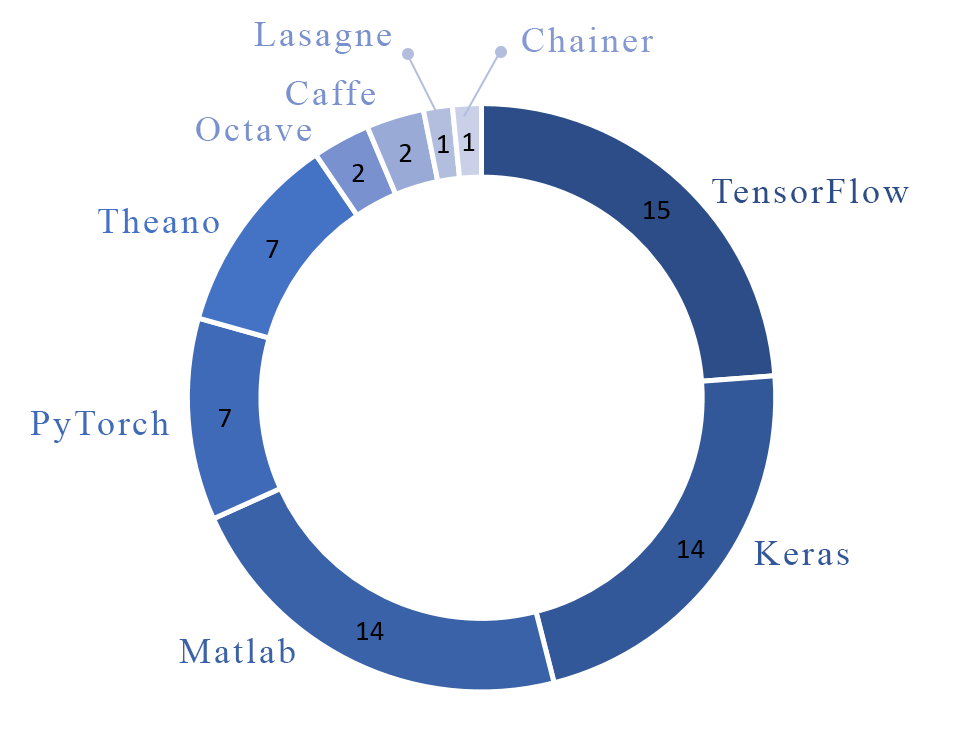}
    
    \caption{Number times each deep learning tool is used for diagnosing epileptic seizures in reviewed papers.}
    \label{fig:one}
\end{figure}

In traditional machine learning algorithms, most simulations were executed in the Matlab software environment, but the deep learning models are usually developed using Python programming language and its numerous open-source toolboxes. This change in the applied programming language helped the researchers to contribute to other works more comfortably and recreate the previously reached results more straightforward. Also created the accessibility of computation resource to everyone thanks to cloud computing, and lastly, made a more convenient path for the creation of application specified hardware for biomedical tasks. Figure \ref{fig:one} shows that, the Tensorflow and one of its high-level APIs, Keras, are widely used for epileptic seizure detection using deep learning in reviewed works due to their versatility and applicability.

% \begin{figure}[t]
%     \centering
%     \includegraphics[width=3.5in ]{pics/figs/chart_drawer_chart4.png}
    
%     \caption{Different deep learning tools and their popularity \cite{sixteen}.}
%     \label{fig:base}
% \end{figure}

\begin{figure}[t]
    \centering
    \includegraphics[width=3.5in]{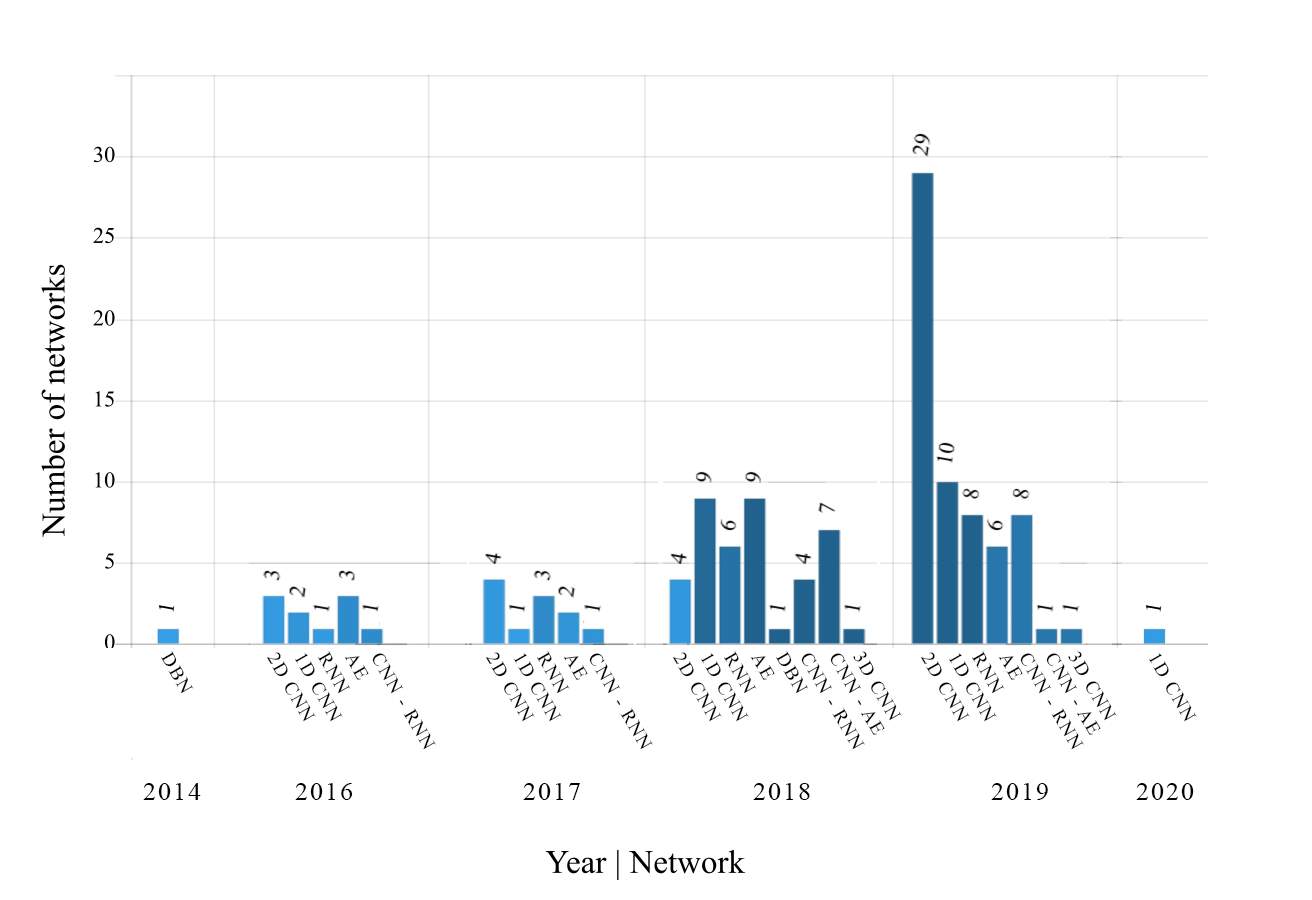}  
    
    \caption{Count of deep learning networks use for epileptic seizure detection in reviewed works.}
    \label{fig:two}
\end{figure}

Since 2016, substantial research has embarked on the field of identifying epilepsy using deep learning models, such as Convolutional Neural Networks (CNN), Recurrent Neural Networks (RNN), Deep Belief Networks (DBN), Autoencoders (AE), CNN-RNN, and CNN-AE \cite{seventeen,eighteen,ninteen,twenty}. The number of studies in this area using deep learning is growing by proposing new efficient models. Figure \ref{fig:two} provides the overview of number of studies conducted using various deep learning models from 2014 to 2020 in detecting epileptic seizures.

The main goals of this study are as follows:
\begin{itemize}
    \item Providing information on  available EEG datasets.
    \item Reviewing works done using various deep learning models for automated detection of epileptic seizures by various modalities.
    \item Introducing future challenges on the detection of epileptic seizures.
    \item Analyzing the best performing model for various modalities of data.
    % \item Reviewing all papers utilizing deep learning based epileptic seizure detection from EEG signal to the best of authors knowledge.
    % \item Investigating works done using other methods of screening for epileptic seizure detection such as MRI.
    % \item Analyzing cloud-based and hardware studies in this field and future directions.
\end{itemize}
Epileptic seizures detection using deep learning is discussed in section II. Section III describes the non-EEG based epileptic seizure detection. Hardware used for epileptic seizures detection is provided in section IV. Discussion on the paper is outlined in section V. The challenges faced by employing deep learning methods for the epileptic seizure detection are summarized in section VI. Finally,  the conclusion and future work are delineated in section VII. 
\section{Epileptic Seizure Detection Based on Deep Learning Techniques}
\begin{figure}[t]
    \centering
    \includegraphics[width=3.5in ]{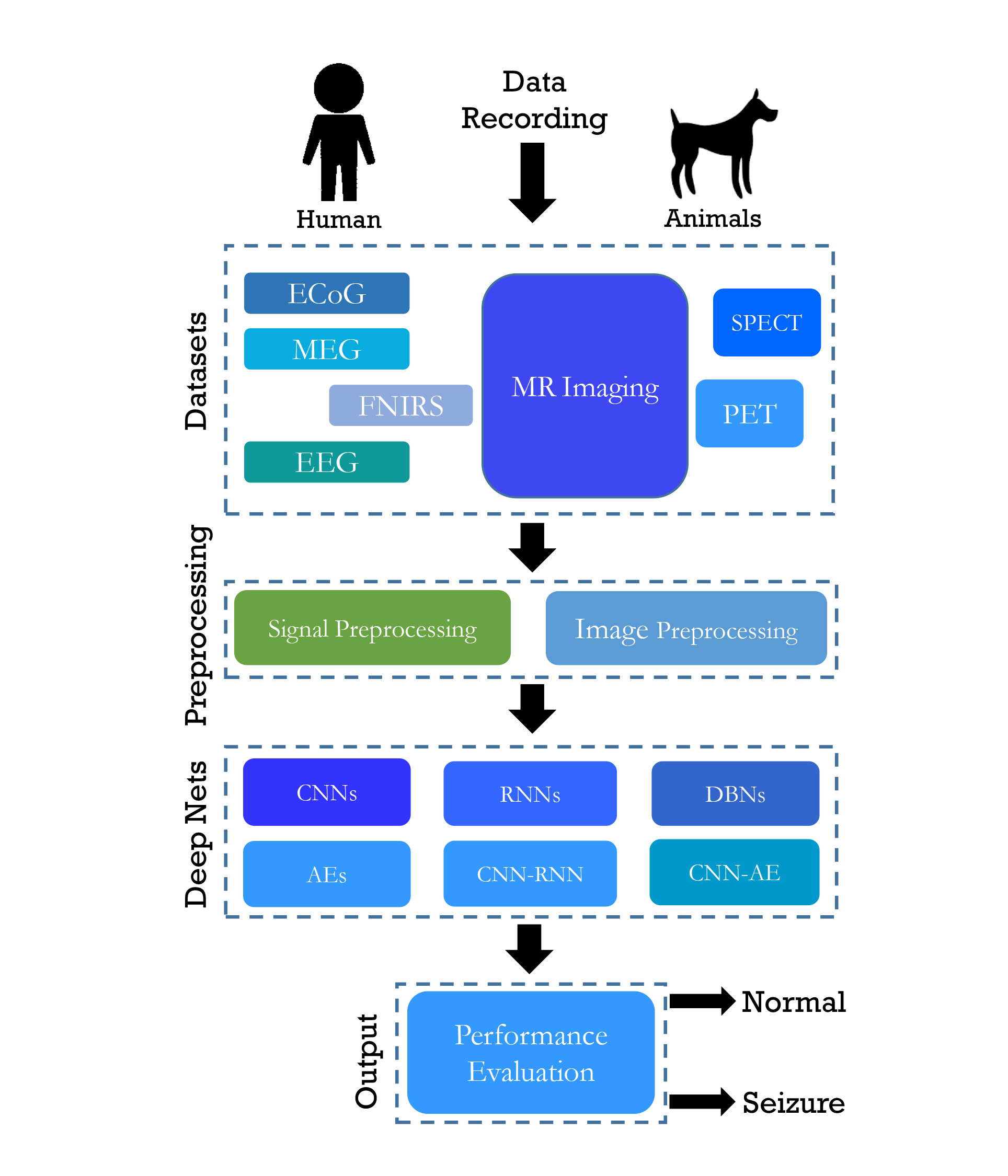}
    
    \caption{Block diagram of a deep learning based CAD system for epileptic seizures.}
    \label{fig:method}
\end{figure}

Figure \ref{fig:method} illustrates the working of a Computer-Aided Diagnosis System (CADS) for epileptic seizures using deep learning methods. The input to the deep learning model can be EEG, MEG, Electrocorticography (ECoG), functional Near-InfraRed Spectroscopy (fNIRS), PET, Single-Photon Emission Computed Tomography
(SPECT), MRI. Then the signal is subjected to the preprocessing to remove the noise. Then these noise eliminated signals are used to develop the deep learning models. The performance of the model is evaluated using accuracy, sensitivity, and specificity. Additionally, a table combining all the works reviewed on epileptic seizure detection using deep learning is presented in the table form in Appendix A of the paper.

\subsection{Epileptic Datasets}
\begin{figure}[t]
    \centering
    \includegraphics[width=3.5in]{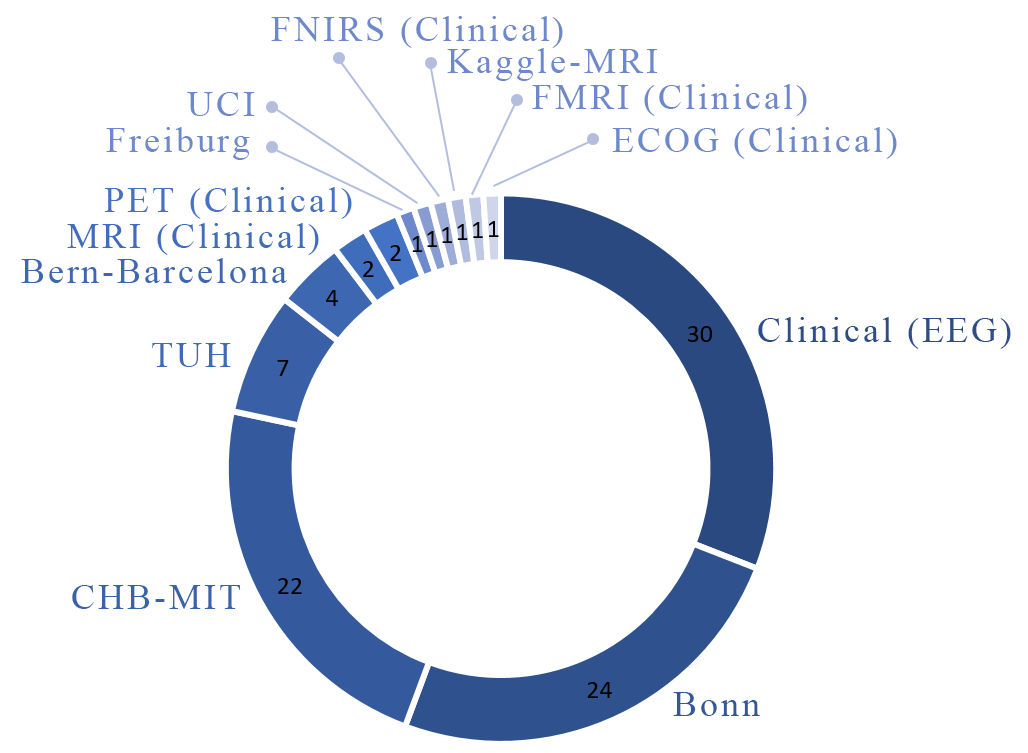}
    
    \caption{Number times each dataset is used to detect epileptic seizures in reviewed works (all clinical dataset usages are summed in related cells).}
    \label{fig:dataset}
\end{figure}

Datasets play an important role in developing accurate and robust CADS. Multiple EEG datasets, namely, Freiburg \cite{twentyOne}, CHB-MIT \cite{twentyTwo}, Kaggle \cite{twentyThree}, Bonn \cite{twentyFour}, Flint-Hills \cite{thirteen}, Bern-Barcelona \cite{twentyFive}, Hauz Khas \cite{thirteen}, and Zenodo \cite{twentySix} are available to develop the automated epileptic seizure detection systems. The signals from these datasets are recorded either intracranially or from the scalp of humans or animals. The supplementary information for each dataset is listed in table \ref{tabledataset}.
\begin{table*}[t]
\caption{List of popular epileptic seizure datasets.}
\label{tabledataset}
  \centering
  \begin{tabular}{|c|c|c|c|c|c|}
  \hline
  \multicolumn{1}{|c|}{\multirow{2}{*}{\shortstack[1]{Dataset}}} & \multicolumn{1}{c|}{\multirow{2}{*}{\shortstack[1]{Number of\\Patients}}} & \multicolumn{1}{c|}{\multirow{2}{*}{\shortstack[1]{Number of\\Seizures}}} & \multicolumn{1}{c|}{\multirow{2}{*}{\shortstack[1]{Recording}}} & \multicolumn{1}{c|}{\multirow{2}{*}{\shortstack[1]{Total\\Duration (hour)}}} & \multicolumn{1}{c|}{\multirow{2}{*}{\shortstack[1]{Sampling\\Frequency (Hz)}}}\\
  \multicolumn{1}{|c|}{} & \multicolumn{1}{c|}{} & \multicolumn{1}{c|}{} & \multicolumn{1}{c|}{} & \multicolumn{1}{c|}{} & \multicolumn{1}{c|}{}\\
  \hline
  
  Flint-Hills \cite{thirteen} & 10 & 59 & Continues intracranial ling term ECoG & 1419 & 249\\
  \hline
  Hauz Khas \cite{thirteen} & 10 & NA & Scalp EEG (sEEG) & NA & 200\\
  \hline
  Freiburg \cite{twentyOne} & 21 & 87 & Intracranial Electroencephalography (IEEG) & 708 & 256 \\
  \hline
  CHB-MIT \cite{twentyTwo} & 22 & 163 & sEEG & 844 & 256\\
  \hline
  \multicolumn{1}{|c|}{\multirow{2}{*}{\shortstack[1]{Kaggle \cite{twentyThree}}}} & 5 dogs & \multicolumn{1}{c|}{\multirow{2}{*}{\shortstack[1]{48}}} & \multicolumn{1}{c|}{\multirow{2}{*}{\shortstack[1]{IEEG}}} &
  \multicolumn{1}{c|}{\multirow{2}{*}{\shortstack[1]{627}}} & 400\\
  \cline{2-2}
  \cline{6-6}
\multicolumn{1}{|c|}{} & 2 patients & \multicolumn{1}{c|}{} & \multicolumn{1}{c|}{} & \multicolumn{1}{c|}{} & 5 KHz\\
\hline
Bonn \cite{twentyFour} & 10 & NA & Surface and IEEG & 39 m & 173.61\\
\hline
Barcelona \cite{twentyFive} & 5 & 3750 & IEEG & 83 & 512\\
\hline
Zenodo \cite{twentySix} & 79 neonatal & 460 & sEEG & 74 m & 256\\
\hline
  \end{tabular}
\end{table*}
Figure \ref{fig:dataset} shows the number of times each dataset is used to detect epileptic seizures in reviewed researches. It is observable that the Bonn dataset is the most employed in reviewed research.
\subsection{Preprocessing}
In a deep learning based CAD system, EEG signal preprocessing commonly involves three steps of noise removal, normalization, and signal preparation for deep learning network applications. In the noise removal step, finite impulse response (FIR) or infinite impulse response (IIR) filters are usually used to eliminate extra signal noise. Normalization is then performed using various schemes such as the z-score technique. Finally, different time domain, frequency, and time-frequency methods are employed to prepare the signals for the deployment of deep networks. 
\subsection{Review of Deep Learning Techniques}
In contrast to conventional neural networks, or so-called shallow networks, deep neural networks are structures with more than two hidden layers. Some recent deep nets have more than hundreds of layers \cite{seventeen}. This increase in the size of the networks results in a massive rise in the number of parameters of the network, requiring appropriate methods for learning, and also measures to avoid overfitting of the learned network. Convolutional networks use filters convolved with input patterns instead of multiplying a weight vector (matrix), which reduces the number of trainable parameters dramatically.

Furthermore, other methods are suggested to help the network to learn, as well \cite{goodfellow}. Pooling layers reduce the size of the input pattern to the next convolutional layer. Batch normalization, dropout, early stopping, unsupervised or semi-unsupervised learning, and regularization techniques prevent the learned network from overfitting and increase the learning ability and speed. The AE and DBN are employed as unsupervised learning and then fine-tuned to avoid overfitting for limited labeled data. Long-Short-Time-Memory (LSTM) and Gated-Recurrent-Units (GRU) are recurrent neural networks capable of revealing the long term time dependencies of data samples. 
\subsubsection{Convolutional Neural Networks (CNNs)}
CNNs are one class of the most popular deep learning networks in which most of the researches in machine learning have been devoted to these networks \cite{seventeen}. They were initially presented for image processing applications but have recently been adopted to one and two-dimensional architectures for diagnosis and prediction of diseases using biological signals \cite{addedOne}. This class of deep learning networks is widely used for the detection of epileptic seizures using EEG signals. In two dimensional convolutional neural networks (2D-CNN), the one dimensional (1D) EEG signals are first transformed into two dimensions employing visualization methods such as spectrogram, higher order bispectrum, and wavelet transforms, then are applied to the input of the convolutional network. In 1D architectures, the EEG signals are applied in the form of one dimensional to the input of convolutional networks. In these networks, changes are made to the core architecture of 2D-CNN that are capable of processing the 1D-EEG signals. Therefore, since both 2D and one-dimensional convolutional neural networks (1D-CNNs) are used in the field of epileptic seizures detection, they are investigated separately.

\textsc{2D - Convolutional Neural Networks}

\begin{figure*}[ht]
    \centering
    \includegraphics[width=7in ]{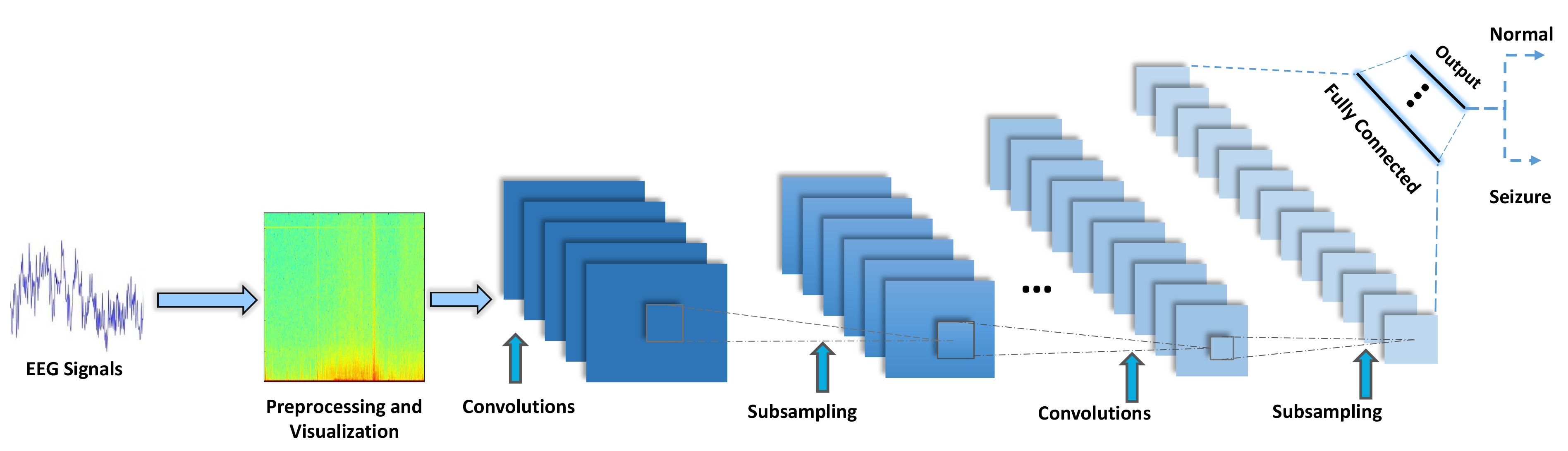}
    
    \caption{A typical 2D-CNN for epileptic seizure detection.}
    \label{fig:cnn2d}
\end{figure*}

Nowadays, deep 2D networks are applied to resolve a wide range of computer vision obstacles such as image segmentation \cite{twentySeven}, medical image classification \cite{twentyEight}, and face recognition \cite{face}. First, in 2012, Krizovsky et al. \cite{alexnet} suggested this network to solve image classification problems, and then quickly use similar networks for all different tasks such as medical image classification, in an effort to obviate the difficulties of previous networks and solve more intricate problems, increased remarkably. Figure \ref{fig:cnn2d} shows a general form of a 2D-CNN used for epileptic seizure detection. The application of 2D-CNN architectures is arguably the most important architecture of deep neural nets. Also, more information about visualization and preprocessing method can be found in Appendix A.

In one study \cite{thirty}, the SeizNet 16-layer convolutional network is introduced, with additional dropout layers and batch normalization (BN) behind each convolutional layer having a structure similar to VGG-Net. The researchers in \cite{thirtyTwo} presented a new 2D-CNN model that can extract the spectral and temporal characteristics of EEG signals and used them to learn the general structure of seizures. Zuo et al. \cite{thirtyThree} developed the diagnosis of Higher Frequency Oscillations (HFO) epilepsy from 16-layer 2D-CNN and EEG signals. A deep learning framework called SeizureNet is proposed in \cite{thirtyFour} using convolution layers with dense connections. A novel deep learning model called the temporal graph convolutional network (TGCN) has been introduced by Covert et al. \cite{thirtySeven}, comprising of five architectures with 14, 18, 22, 23 and 26 layers. Bouaziz et al. \cite{fourty} split the EEG signals of CHB-MIT with 23 channels into 2-second time windows and then converted them into density images (spatial representation), which were fed as inputs to the CNN network.

\textsc{Alexnet}

FeiFei Li, Professor of Stanford University, created a dataset of labeled images of real-world objects and termed her project as ImageNet\cite{imagenet}. ImageNet organized a computer vision competition called ILSVRC annually to solve the image classification problems. Alex Krizhevsky revolutionized the image classification world with his algorithm, AlexNet, which won the 2012 ImageNet challenge and started the whole deep learning era \cite{alexnet}. AlexNet won the competition by achieving the top-5 test accuracy of 84.6\%. Taqi et al. \cite{fourtyTwo} used the AlexNet network to diagnose focal epileptic seizures. This proposed network used the feature extraction approach and eventually applied the softmax layer for classification purposes and achieved 100\% accuracy. In another research, the AlexNet network was employed \cite{fourtyThree}. They transformed the 1D signal in to 2D image by passing through Signal2Image (S2I) module. The several methods used in this are signal as image, spectrogram, one layer 1D-CNN, and two-layer 1D-CNN.  

\textsc{VGG}

A research team at Oxford proposed the Visual Geometry Group (VGG) CNN model in 2014 \cite{vgg}. They configured various models, and one such model was called VGG-16 was submitted to ILSVRC 2014 competition. The model was known as VCG-16 because it comprised of 16 layers. It delivered an excellent performance in image detection and classification problems. Ahmedt-Aristizabal et al. \cite{fourtyFour} performed VGG-16 architecture to diagnose epilepsy from facial images. Their proposed approach attempted to extract and classify semiological patterns of facial states automatically. After recording the images, the proposed VGG architecture is trained primarily by well-known datasets, followed by various networks such as 1D-CNN and LSTM in the last few layers. In \cite{fourtyThree}, the VGG network used one-dimensional and two-dimensional signals. To train the models, Adam's optimizer and a cross-entropy error function were used. They used the batch size and number of epochs as 20 and 100 respectively. The idea of detecting epileptic seizures on the sEEG signal plots was examined by Emami et al. \cite{fourtyFive}. In the pre-processing step, the signals were segmented into different time windows and VGG-16 was used for classification using small (3×3) convolution filters to efficiently detect small EEG signal changes. This architecture was pre-trained by applying an ImageNet dataset to differentiate 1000 classes, and the last two layers had 4096 and 1000 dimensional vectors. They modified these last two layers to have 32 and 2 dimensions, respectively, to detect seizure and non-seizure classes. 

\textsc{GoogLeNet}

GoogLeNet won the 2014 ImageNet competition with 93.3\% top-5 test accuracy \cite{googlenet}. This 22-layer network was called GoogLeNet to honor Yann Lecun, who designed LeNet. Before the introduction of GoogLeNet, it was stated that by going deep, one could achieve better accuracy and results. Nevertheless, the google team proposed an architecture called inception, which achieved better performance by not going deep but by better design. It represented a robust design by using filters of different sizes on the same image. In the field of EEG signal processing to diagnose epileptic seizures, this architecture has recently received the attention of researchers. Taqi et al. \cite{fourtyTwo} used this network in their preliminary researches to diagnose epileptic seizures. Their model was used to extract features from the Bern-Barcelona dataset and achieved excellent results.

\textsc{ResNet}

\begin{figure}[t]
    \centering
    \includegraphics[width=3.5in]{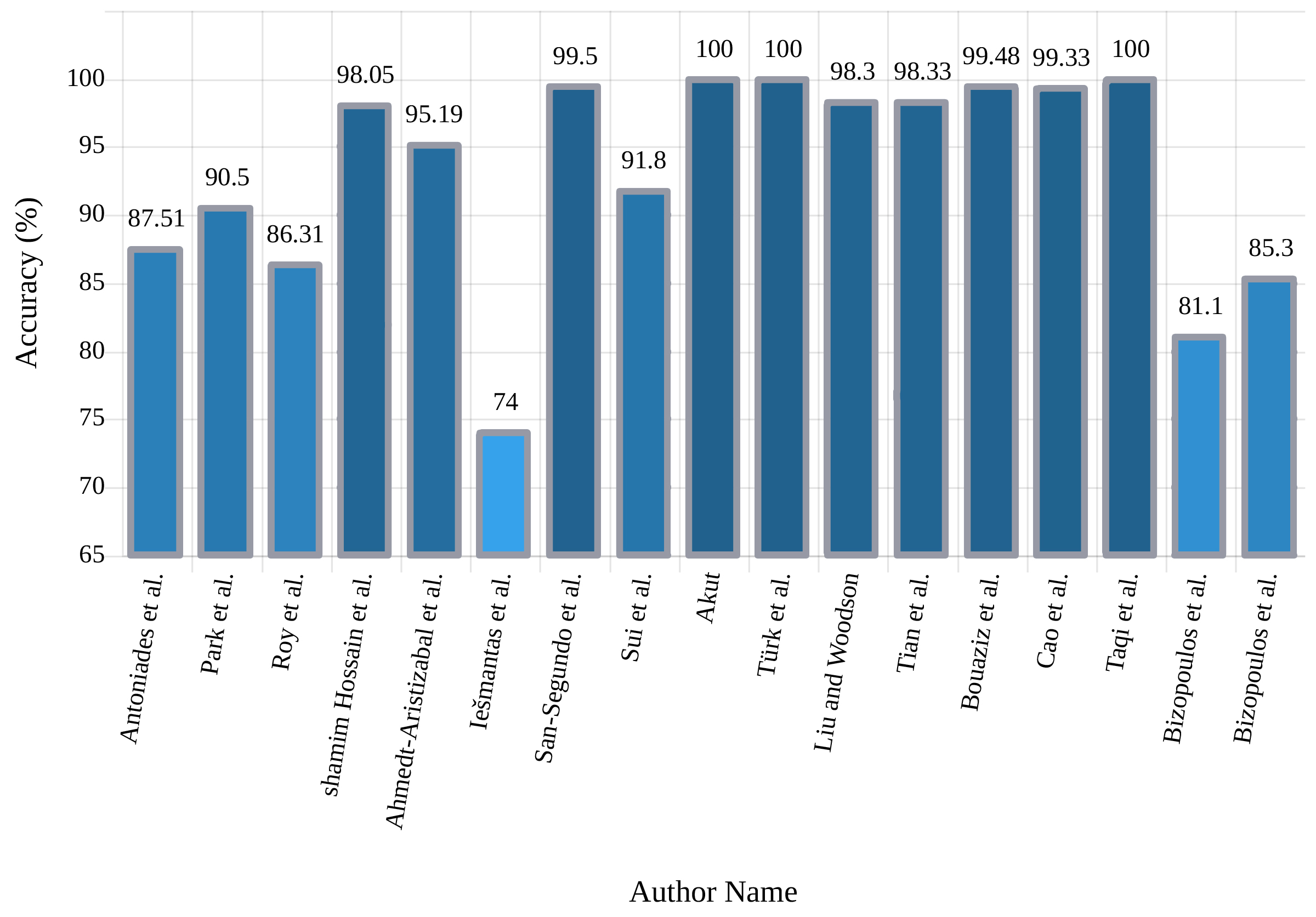}
    
    \caption{Sketch of accuracy (\%) versus authors obtained using 2D-CNN models for seizure detection.}
    \label{fig:cnn2dchart}
\end{figure}

Microsoft's ResNet won ImageNet challenge with 96.4\% accuracy by applying a 152-layer network which utilized a Resnet module \cite{resnet}. In this network, residual blocks were introduced, which were capable of training deep architecture by using identity skip connections which copied inputs of each layer to the next layer. The idea was to learn something different and new in the next layer. So far, not much research has been accomplished on the implementation of ResNet networks to diagnose epilepsy, but may grow significantly in coming days. Bizopoulos et al. \cite{fourtyThree} introduced two ResNet and DenseNet architectures to diagnose epileptic seizures and attained good results. They showed that S2I-DenseNet base model with an average of 70 epochs was sufficient to gain the best accuracy of 85.3\%. A summary of the research accomplished in this section is reported in Table \ref{tablecnn2d}. A comparison of accuracy obtained by each study is shown in Figure \ref{fig:cnn2dchart}.

\begin{table}[t]
\caption{Summary of related works done using 2D CNNs.}
\label{tablecnn2d}
\centering
\resizebox{3.5in}{!}{
\begin{tabular}{|c|c|c|c|c|}
\hline
\multicolumn{1}{|c|}{\multirow{2}{*}{Work}} & \multicolumn{1}{c|}{\multirow{2}{*}{Networks}} & \multicolumn{1}{c|}{\multirow{2}{*}{\shortstack[1]{Number \\of Layers}}} & \multicolumn{1}{c|}{\multirow{2}{*}{Classifier}} & \multicolumn{1}{c|}{\multirow{2}{*}{Accuracy\%}}\\
\multicolumn{1}{|c|}{} & \multicolumn{1}{c|}{} & \multicolumn{1}{c|}{} & \multicolumn{1}{c|}{} &  \multicolumn{1}{c|}{}\\
\hline
\multicolumn{1}{|c|}{\multirow{2}{*}{\shortstack[1]{\cite{twentyNine}}}} & \multicolumn{1}{c|}{\multirow{2}{*}{\shortstack[1]{2D-CNN}}} & 3 & \multicolumn{1}{c|}{\multirow{2}{*}{\shortstack[1]{Logistic Regression (LR)}}} &  \multicolumn{1}{c|}{\multirow{2}{*}{\shortstack[1]{87.51}}}\\
\cline{3-3}
\multicolumn{1}{|c|}{} & \multicolumn{1}{c|}{} & 4 & \multicolumn{1}{c|}{} &  \multicolumn{1}{c|}{}\\
\hline
% \cite{sixty} & 2D-CNN & 12 & softmax & NA\\
% \hline
\cite{sixtyOne} & 2D-CNN & 9 & softmax & NA\\
\hline
\multicolumn{1}{|c|}{\multirow{2}{*}{\shortstack[1]{\cite{sixtyTwo}}}} & \multicolumn{1}{c|}{\multirow{2}{*}{\shortstack[1]{Combination of 1D-\\CNN and 2D-CNN}}} & \multicolumn{1}{c|}{\multirow{2}{*}{\shortstack[1]{11}}} & \multicolumn{1}{c|}{\multirow{2}{*}{\shortstack[1]{sigmoid}}} &  \multicolumn{1}{c|}{\multirow{2}{*}{\shortstack[1]{90.58}}}\\
\multicolumn{1}{|c|}{} & \multicolumn{1}{c|}{} & \multicolumn{1}{c|}{} & \multicolumn{1}{c|}{} &  \multicolumn{1}{c|}{}\\
\hline
% \cite{sixtyThree} & 2D-CNN & 13 & softmax & NA \\
% \hline
\cite{sixtyFour} & 2D-CNN & 18 & softmax & NA \\
\hline
\cite{sixtyFive} & 2D-CNN/MLP hybrid & 11 & sigmoid & NA\\
\hline
\cite{fourtySix} & 2D-CNN & 9 & softmax & 86.31 \\
\hline
\cite{thirty} & SeizNet & 16 & NA & NA \\
\hline
\multicolumn{1}{|c|}{\multirow{2}{*}{\shortstack[1]{\cite{thirtyOne}}}} & \multicolumn{1}{c|}{\multirow{2}{*}{\shortstack[1]{2D-CNN with\\1D-CNN}}} & \multicolumn{1}{c|}{\multirow{2}{*}{\shortstack[1]{12}}} & \multicolumn{1}{c|}{\multirow{2}{*}{\shortstack[1]{softmax}}} &  \multicolumn{1}{c|}{\multirow{2}{*}{\shortstack[1]{NA}}}\\
\multicolumn{1}{|c|}{} & \multicolumn{1}{c|}{} & \multicolumn{1}{c|}{} & \multicolumn{1}{c|}{} &  \multicolumn{1}{c|}{}\\
\hline
\cite{thirtyTwo} & 2D-CNN & 9 & softmax & 98.05\\
\hline
\cite{thirtyThree} & 2D-CNN & 16 & softmax & NA\\
\hline
\cite{thirtyFour} & SeizureNet & 133 & softmax & NA\\
\hline
\cite{fourtyFour} & 2D-CNN & VGG-16,8 & SVM & 95.19\\
\hline
% \cite{sixtySix} & 2D-CNN & 8 & softmax, SVM & 86.25\\
% \hline
\cite{thirtyFive} & 2D-CNN & 6 & softmax & 74\\
\hline
% \multicolumn{1}{|c|}{\multirow{2}{*}{\shortstack[1]{\cite{sixtySeven}}}} & \multicolumn{1}{c|}{\multirow{2}{*}{\shortstack[1]{2D-CNN(similar\\to LeNet5)}}} & \multicolumn{1}{c|}{\multirow{2}{*}{\shortstack[1]{6}}} & \multicolumn{1}{c|}{\multirow{2}{*}{\shortstack[1]{BLDA}}} &  \multicolumn{1}{c|}{\multirow{2}{*}{\shortstack[1]{84.92}}}\\
% \multicolumn{1}{|c|}{} & \multicolumn{1}{c|}{} & \multicolumn{1}{c|}{} & \multicolumn{1}{c|}{} &  \multicolumn{1}{c|}{}\\
% \hline
\multicolumn{1}{|c|}{\multirow{2}{*}{\shortstack[1]{\cite{sixtyEight}}}} & \multicolumn{1}{c|}{\multirow{2}{*}{\shortstack[1]{2D-CNN}}} & \multicolumn{1}{c|}{\multirow{2}{*}{\shortstack[1]{12}}} & \multicolumn{1}{c|}{\multirow{2}{*}{\shortstack[1]{softmax and\\sigmoid}}} &  \multicolumn{1}{c|}{\multirow{2}{*}{\shortstack[1]{99.50}}}\\
\multicolumn{1}{|c|}{} & \multicolumn{1}{c|}{} & \multicolumn{1}{c|}{} & \multicolumn{1}{c|}{} &  \multicolumn{1}{c|}{}\\
\hline
\cite{thirtySix} & 2D-CNN & 16 & softmax & 91.80\\
\hline
\multicolumn{1}{|c|}{\multirow{5}{*}{\shortstack[1]{\cite{thirtySeven}}}} & \multicolumn{1}{c|}{\multirow{5}{*}{\shortstack[1]{TGCN}}} & 14 & \multicolumn{1}{c|}{\multirow{5}{*}{\shortstack[1]{sigmoid}}} &  \multicolumn{1}{c|}{\multirow{5}{*}{\shortstack[1]{NA}}}\\
\cline{3-3}
\multicolumn{1}{|c|}{} & \multicolumn{1}{c|}{} & 18 & \multicolumn{1}{c|}{} &  \multicolumn{1}{c|}{}\\
\cline{3-3}
\multicolumn{1}{|c|}{} & \multicolumn{1}{c|}{} & 22 & \multicolumn{1}{c|}{} &  \multicolumn{1}{c|}{}\\
\cline{3-3}
\multicolumn{1}{|c|}{} & \multicolumn{1}{c|}{} & 22 & \multicolumn{1}{c|}{} &  \multicolumn{1}{c|}{}\\
\cline{3-3}
\multicolumn{1}{|c|}{} & \multicolumn{1}{c|}{} & 26 & \multicolumn{1}{c|}{} &  \multicolumn{1}{c|}{}\\
\hline
\cite{thirtyEight} & 2D-CNN & 23 & softmax & 100\\
\hline
\cite{sixtyNine} & 2D-CNN & 5 & softmax & 100\\
\hline
\multicolumn{1}{|c|}{\multirow{4}{*}{\shortstack[1]{\cite{thirtyNine}}}} & \multicolumn{1}{c|}{\multirow{4}{*}{\shortstack[1]{2D-CNN}}} & \multicolumn{1}{c|}{\multirow{4}{*}{\shortstack[1]{14}}} & \multicolumn{1}{c|}{\multirow{4}{*}{\shortstack[1]{softmax}}} &  \multicolumn{1}{c|}{\multirow{2}{*}{\shortstack[1]{2 classes\\98.30}}}\\
\multicolumn{1}{|c|}{} & \multicolumn{1}{c|}{} & \multicolumn{1}{c|}{} & \multicolumn{1}{c|}{} &  \multicolumn{1}{c|}{}\\
\cline{5-5}
\multicolumn{1}{|c|}{} & \multicolumn{1}{c|}{} & \multicolumn{1}{c|}{} & \multicolumn{1}{c|}{} &  \multicolumn{1}{c|}{\multirow{2}{*}{\shortstack[1]{3 classes\\90.10}}}\\
\multicolumn{1}{|c|}{} & \multicolumn{1}{c|}{} & \multicolumn{1}{c|}{} & \multicolumn{1}{c|}{} &  \multicolumn{1}{c|}{}\\
\hline
\multicolumn{1}{|c|}{\multirow{3}{*}{\shortstack[1]{\cite{seventy}}}} & \multicolumn{1}{c|}{\multirow{2}{*}{\shortstack[1]{2D-CNN}}} & 7 & \multicolumn{1}{c|}{\multirow{3}{*}{\shortstack[1]{MV-TSK-FS}}} &  \multicolumn{1}{c|}{\multirow{3}{*}{\shortstack[1]{98.33}}}\\
\cline{3-3}
\multicolumn{1}{|c|}{} & \multicolumn{1}{c|}{} & 5 & \multicolumn{1}{c|}{} &  \multicolumn{1}{c|}{}\\
\cline{2-3}
\multicolumn{1}{|c|}{} & 3D-CNN & 8 & \multicolumn{1}{c|}{} &  \multicolumn{1}{c|}{}\\
\hline
\cite{fourty} & 2D-CNN & 8 & softmax & 99.48\\
\hline
% \cite{seventyOne} & 2D-CNN & 8 & MLP & 94.10\\
% \hline
\multicolumn{1}{|c|}{\multirow{2}{*}{\shortstack[1]{\cite{fourtyOne}}}} & \multicolumn{1}{c|}{\multirow{2}{*}{\shortstack[1]{2D-CNN}}} & 23 & sigmoid &  \multicolumn{1}{c|}{\multirow{2}{*}{\shortstack[1]{NA}}}\\
\cline{3-4}
\multicolumn{1}{|c|}{} & \multicolumn{1}{c|}{} & 18 & RF &  \multicolumn{1}{c|}{}\\
\hline
\cite{seventyTwo} & 2D-CNN & 7 & KELM & 99.33\\
\hline
\multicolumn{1}{|c|}{\multirow{3}{*}{\shortstack[1]{\cite{fourtyTwo}}}} & GoogleNet & \multicolumn{1}{c|}{\multirow{3}{*}{\shortstack[1]{Standard\\Networks}}} & \multicolumn{1}{c|}{\multirow{3}{*}{\shortstack[1]{softmax}}} &  \multicolumn{1}{c|}{\multirow{3}{*}{\shortstack[1]{100}}}\\
\cline{2-2}
\multicolumn{1}{|c|}{} & AlexNet & \multicolumn{1}{c|}{} & \multicolumn{1}{c|}{} &  \multicolumn{1}{c|}{}\\
\cline{2-2}
\multicolumn{1}{|c|}{} & LeNet & \multicolumn{1}{c|}{} & \multicolumn{1}{c|}{} &  \multicolumn{1}{c|}{}\\
\hline
\cite{fourtyFive} & 2D-CNN & VGG-16 & softmax & NA\\
\hline
\multicolumn{1}{|c|}{\cite{fourtyThree}} & \multicolumn{2}{c|}{Standard Networks} & softmax & 85.30\\
\hline
\end{tabular}}
\end{table} 
    
\textsc{1D - Convolutional Neural Networks}

\begin{figure*}[t]
    \centering
    \includegraphics[width=7in ]{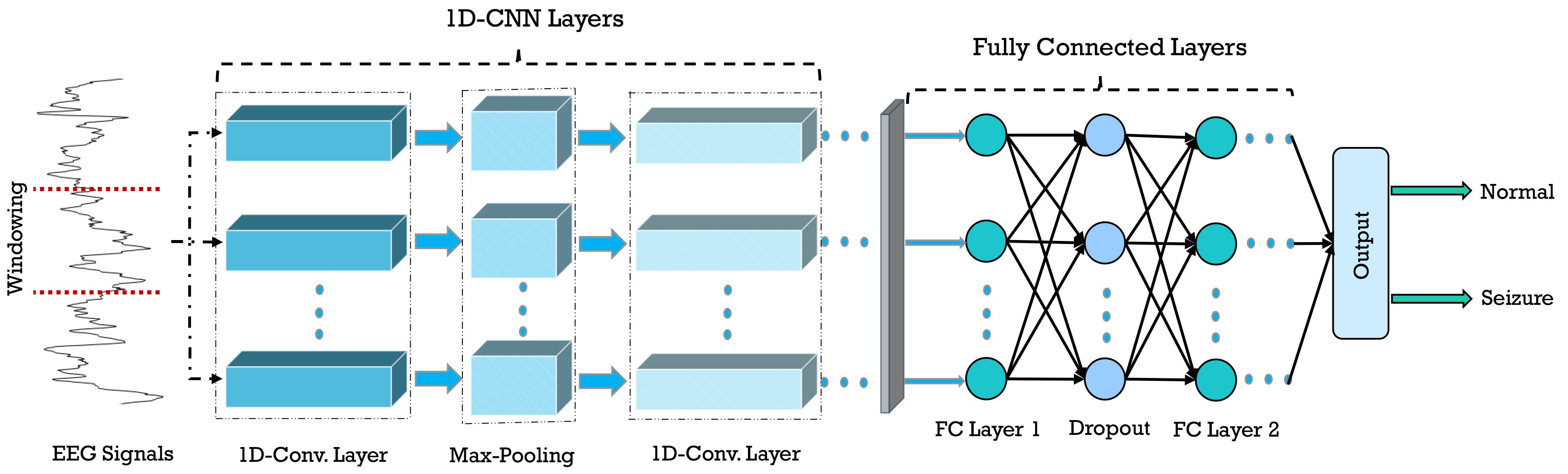}
    
    \caption{Typical sketch of the 1D-CNN model for epileptic seizure detection.}
    \label{fig:cnn1d}
\end{figure*}

1D-CNNs are intrinsically suitable for processing of biological signals such as EEG for detection of epileptic seizures. These architectures present a more straightforward structure and a single pass of them is faster as compared to CNN with 2D architecture, due to lower number of parameters. The most important superiority of 1D to 2D architectures is the possibility of employing pooling and convolutional layers with a larger size. In addition to that, signals are 1D in nature, and using pre-processing methods to transform them to 2D may lead to information loss, yet all the data are preserved in 1D representation. Figure \ref{fig:cnn1d} shows a general form of a 1D-CNN used for epileptic seizure detection.

\begin{figure}[t]
    \centering
    \includegraphics[width=3.5in]{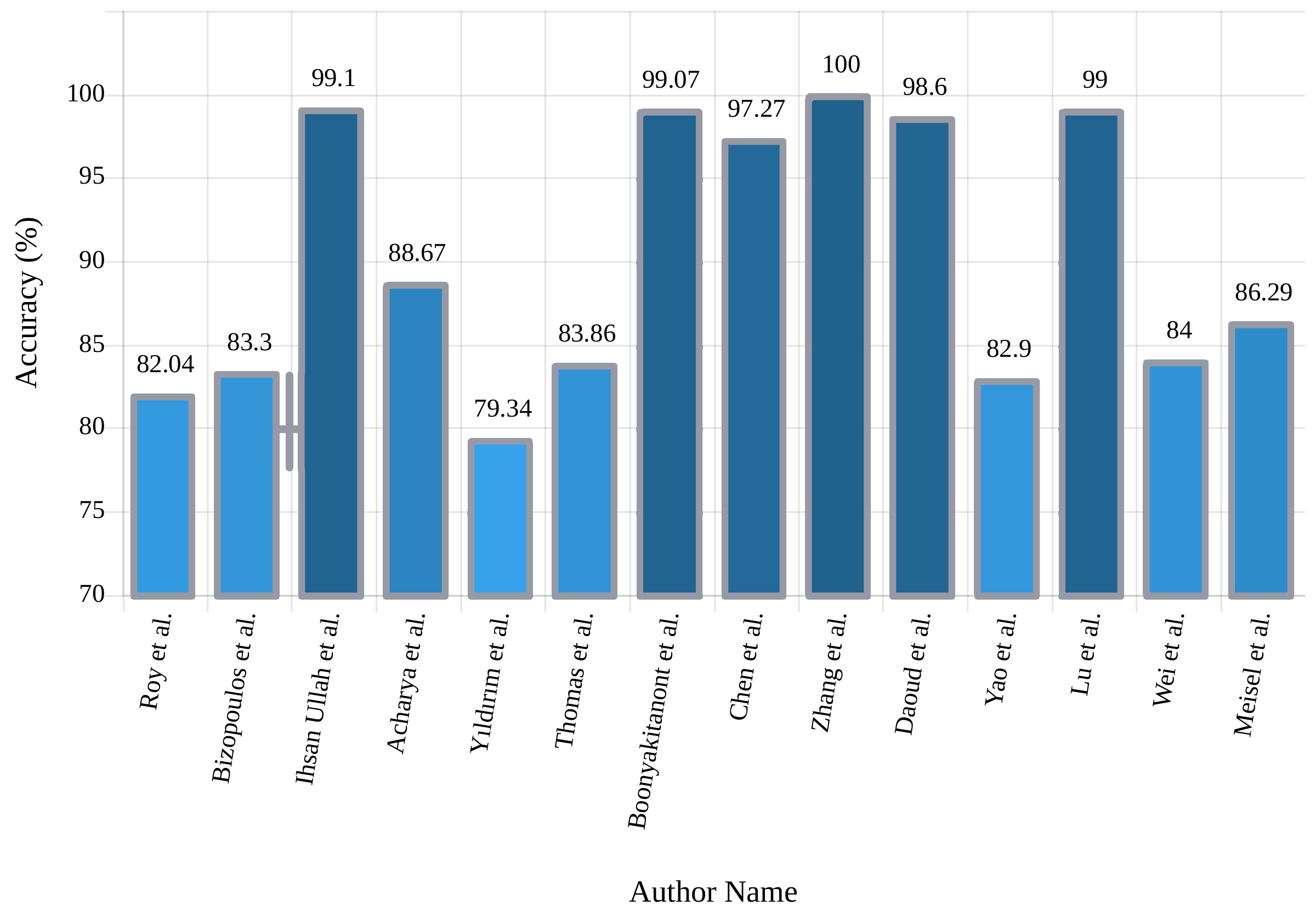}
    
    \caption{Sketch of accuracy (\%) versus authors obtained using 1D-CNN models for seizure detection.}
    \label{fig:cnn1dchart}
\end{figure}

The authors in \cite{fourtyThree} pursued their experiments using one-dimensional LeNet, AlexNet, VGGnet, ResNet, and DenseNet architectures, and applied well-known 2D architectures in 1D space is the first study in this section. In \cite{fourtyNine}, 1D-CNN was used for the feature extraction procedure. The researchers \cite{fifty} used 1D-CNN in another research. In this work, the CHB-MIT dataset is adopted, and the signals from each channel are segmented into 4-second intervals; additionally, overlapping segments are also accepted to increase the data and accuracy. Combining CNNs with traditional feature extraction methods together was explored in \cite{fiftyThree}; they used the Empirical Mode Decomposition (EMD) method for feature extraction, and a CNN was applied to acquire high accuracy in the multi-class classification tasks. In \cite{fiftyFive}, an integrated framework for the diagnosis of epileptic seizures is presented that combines the capability of interpreting probabilistic graphical models (PGMs) with advances in deep learning. The authors in \cite{fiftyEight} submitted a 1D-CNN architecture defined CNN-BP (stading for CNN bipolar). In this work, they used the data from patients monitored with combined foramen ovale (FO) electrodes and EEG surface electrodes. A new scheme to classify EEG signals based on temporal convolution neural networks (TCNN) is introduced by Zhang et al. \cite{fiftyTwo}. Table \ref{tablecnn1d} shows the summary of related works done using 1D CNNs. Figure \ref{fig:cnn1dchart} shows the sketch of accuracy (\%) obtained by various  authors using 1D-CNN models for seizure detection. 

\begin{table}[t]
\caption{Summary of related works done using 1D CNNs.}
\label{tablecnn1d}
\centering
\resizebox{3.5in}{!}{
\begin{tabular}{|c|c|c|c|c|}
\hline
\multicolumn{1}{|c|}{\multirow{2}{*}{Work}} & \multicolumn{1}{c|}{\multirow{2}{*}{Networks}} & \multicolumn{1}{c|}{\multirow{2}{*}{\shortstack[1]{Number \\of Layers}}} & \multicolumn{1}{c|}{\multirow{2}{*}{Classifier}} & \multicolumn{1}{c|}{\multirow{2}{*}{Accuracy\%}}\\
\multicolumn{1}{|c|}{} & \multicolumn{1}{c|}{} & \multicolumn{1}{c|}{} & \multicolumn{1}{c|}{} &  \multicolumn{1}{c|}{}\\
\hline
\cite{fourtySix} & 1D-CNN & 7 & softmax & 82.04\\
\hline
\multicolumn{1}{|c|}{\multirow{3}{*}{\shortstack[1]{\cite{fourtyThree}}}} & \multicolumn{1}{c|}{\multirow{3}{*}{\shortstack[1]{1D-CNN}}} & \multicolumn{2}{c|}{VGGNet - 13} &  \multicolumn{1}{c|}{\multirow{3}{*}{\shortstack[1]{83.30}}}\\
\cline{3-4}
\multicolumn{1}{|c|}{} & \multicolumn{1}{c|}{} & \multicolumn{2}{c|}{VGGNet - 19}  &  \multicolumn{1}{c|}{}\\
\cline{3-4}
\multicolumn{1}{|c|}{} & \multicolumn{1}{c|}{} & \multicolumn{2}{c|}{Densenet - 161}  &  \multicolumn{1}{c|}{}\\
\hline
\cite{seventyThree} & P-1D-CNN & 14 & softmax & 99.10 \\
\hline
\cite{fourtySeven} & 1D-CNN & 13 & softmax & 88.67 \\
\hline
\cite{seventyFour} & MPCNN & 11 & softmax & NA \\
\hline
\cite{fourtyEight} & 1D-FCNN & 11 & softmax & NA \\
\hline
\cite{seventyFive} & 1D-CNN & 5 & binary LR & NA \\
\hline
\cite{seventySix} & 1D-CNN & 23 & softmax & 79.34 \\
\hline
\cite{fourtyNine} & 1D-CNN & 5 & softmax, SVM & 83.86\\
\hline
% \multicolumn{1}{|c|}{\multirow{2}{*}{\shortstack[1]{\cite{seventySeven}}}} & \multicolumn{1}{c|}{\multirow{2}{*}{\shortstack[1]{1D-CNN}}} & 4 & \multicolumn{1}{c|}{\multirow{2}{*}{\shortstack[1]{softmax, LR}}} &  \multicolumn{1}{c|}{\multirow{2}{*}{\shortstack[1]{NA}}}\\
% \cline{3-3}
% \multicolumn{1}{|c|}{} & \multicolumn{1}{c|}{} & 5 & \multicolumn{1}{c|}{} &  \multicolumn{1}{c|}{}\\ 
% \hline
\cite{fifty} & 1D-CNN & 33 & NA & 99.07 \\
\hline
\cite{fiftyOne} & 1D-CNN & 4 & sigmoid & 97.27 \\
\hline
\cite{fiftyTwo} & 1D-TCNN & NA & NA & 100\\
\hline
\cite{fiftyThree} & 1D-CNN & 12 & softmax & 98.60\\
\hline
\cite{fiftyFour} & 1D-CNN & 13 & NA & 82.90 \\
\hline
\multicolumn{1}{|c|}{\multirow{2}{*}{\shortstack[1]{\cite{seventyEight}}}} & \multicolumn{1}{c|}{\multirow{2}{*}{\shortstack[1]{1D-CNN with\\residual connections}}} & \multicolumn{1}{c|}{\multirow{2}{*}{\shortstack[1]{17}}} & \multicolumn{1}{c|}{\multirow{2}{*}{\shortstack[1]{softmax}}} &  99.00\\
\cline{5-5}
\multicolumn{1}{|c|}{} & \multicolumn{1}{c|}{} & \multicolumn{1}{c|}{} & \multicolumn{1}{c|}{} &  91.80\\
\hline
\cite{fiftyFive} & PGM-CNN & 10 & softmax & NA \\
\hline
\cite{seventyNine} & 1D-CNN & 15 & softmax & 84 \\
\hline
\cite{fiftySix} & 1D-CNN & 10 & sigmoid & 86.29 \\
\hline
\cite{fiftySeven} & 1D-CNN & 13 & softmax & NA \\
\hline
\cite{fiftyEight} & 1D-CNN-BP & 14 & sigmoid & NA\\
\hline
\cite{fiftyNine} & 1D-CNN & 9 & sigmoid & NA\\
\hline

\end{tabular}}
\end{table} 

\subsubsection{Recurrent Neural Networks (RNNs)}

\begin{figure}[ht]
    \centering
    \includegraphics[width=3.5in ]{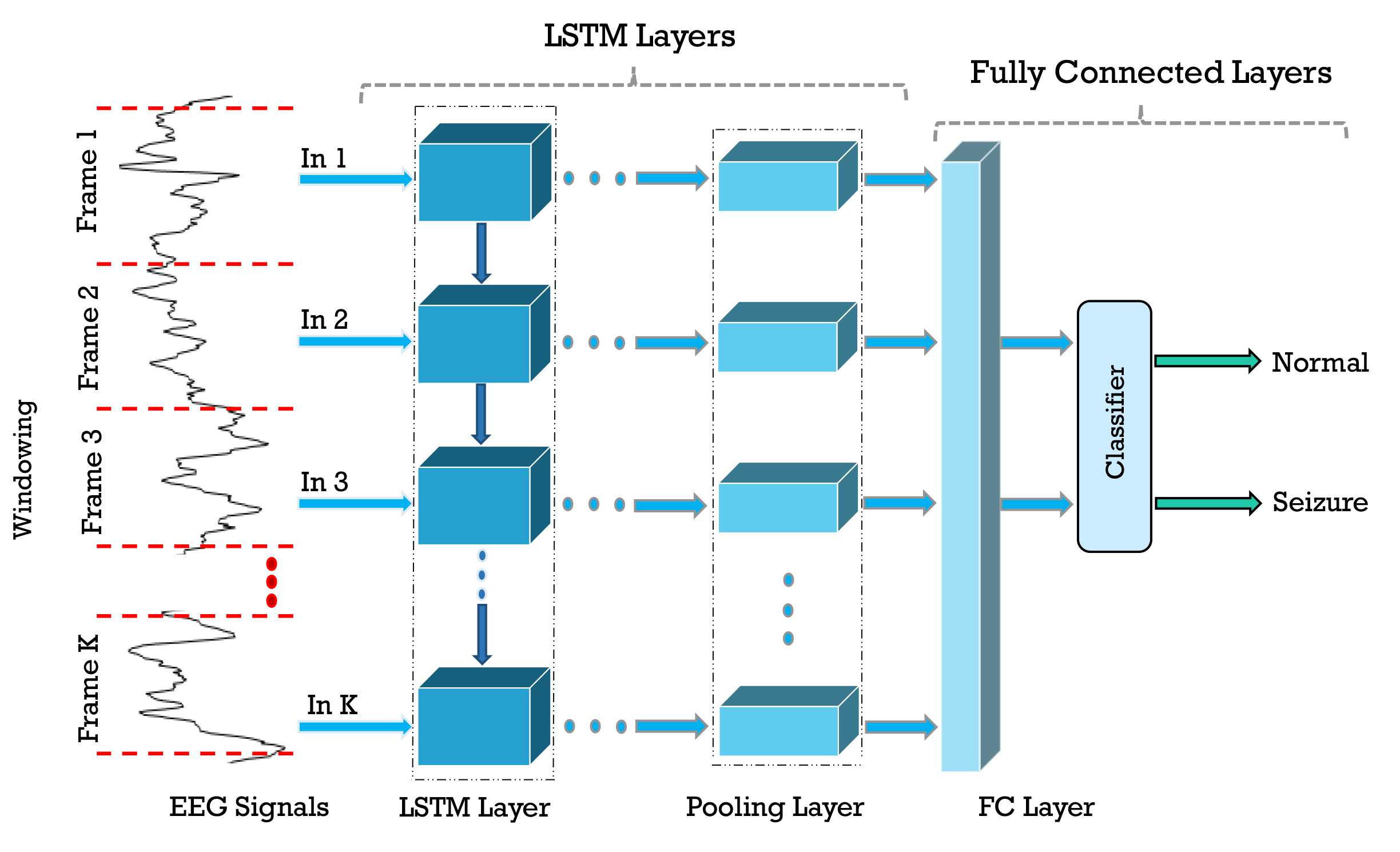}
    
    \caption{Sample RNN model which may be used for seizure detection.}
    \label{fig:lstm}
\end{figure}

Sequential data such as text, signals, and videos, show characteristics like variable and great length, which makes them not suitable for simple deep learning methods \cite{goodfellow}. However, these data form a significant part of the information in the world, compelling the need for deep learning based schemes to process these types of data. RNNs are the solution suggested to overcome the mentioned challenges, and they are widely used for biomedical signal processing. Figure \ref{fig:lstm} shows a general form of RNN used for epileptic seizure detection. In the following section, an overview of popular RNN block structures is presented in addition to the reviewed papers.
 
\textsc{Long Short-Term Memory (LSTM)}

The main problem of a simple recurrent neural network is short-term memory. RNN may leave out key information as it has a hard time transporting information from earlier time steps to the next steps in long sequence. Another drawback of RNN is the vanishing gradient problem \cite{seventeen,eighteen,ninteen,twenty}. The problem arises because of the shrinking of gradients as it back-propagates. To solve the short-term memory problem, LSTM gates were created \cite{seventeen}. The flow of information can be regulated through gates. The gates can preserve the long sequence of necessary data, and throw away the undesired ones. The building block of LSTM is the cell state and its gates. 

In this section, Golmohammadi et al. \cite{sixtyFive} evaluated two LSTM architectures with 3 and 4 layers together with the softmax classifier in their investigation and fetched up satisfactory results. In \cite{fiftyOne}, a 3-layer LSTM deep network is used for feature extraction and classification. The last layer of this network is a sigmoid classification algorithm, and they achieved 96.82\% accuracy. According to directed experiments in \cite{fiftyNine}, they employed two LSTM and GRU architectures. The LSTM, GRU model architecture, comprised a layer of Reshape, four layers of LSTM / GRU with the activator, and one layer of Fully Connected (FC) with sigmoid activator. In another study, Yao et al. \cite{eighty} practiced ten different and ameliorated Independently Recurrent Neural Network (IndRNN) architectures and achieved the best accuracy using Dense IndRNN with attention (DIndRNN) with 31 layers. 

\textsc{Gated Recurrent Unit (GRU)}

\begin{figure}[t]
    \centering
    \includegraphics[width=3.5in]{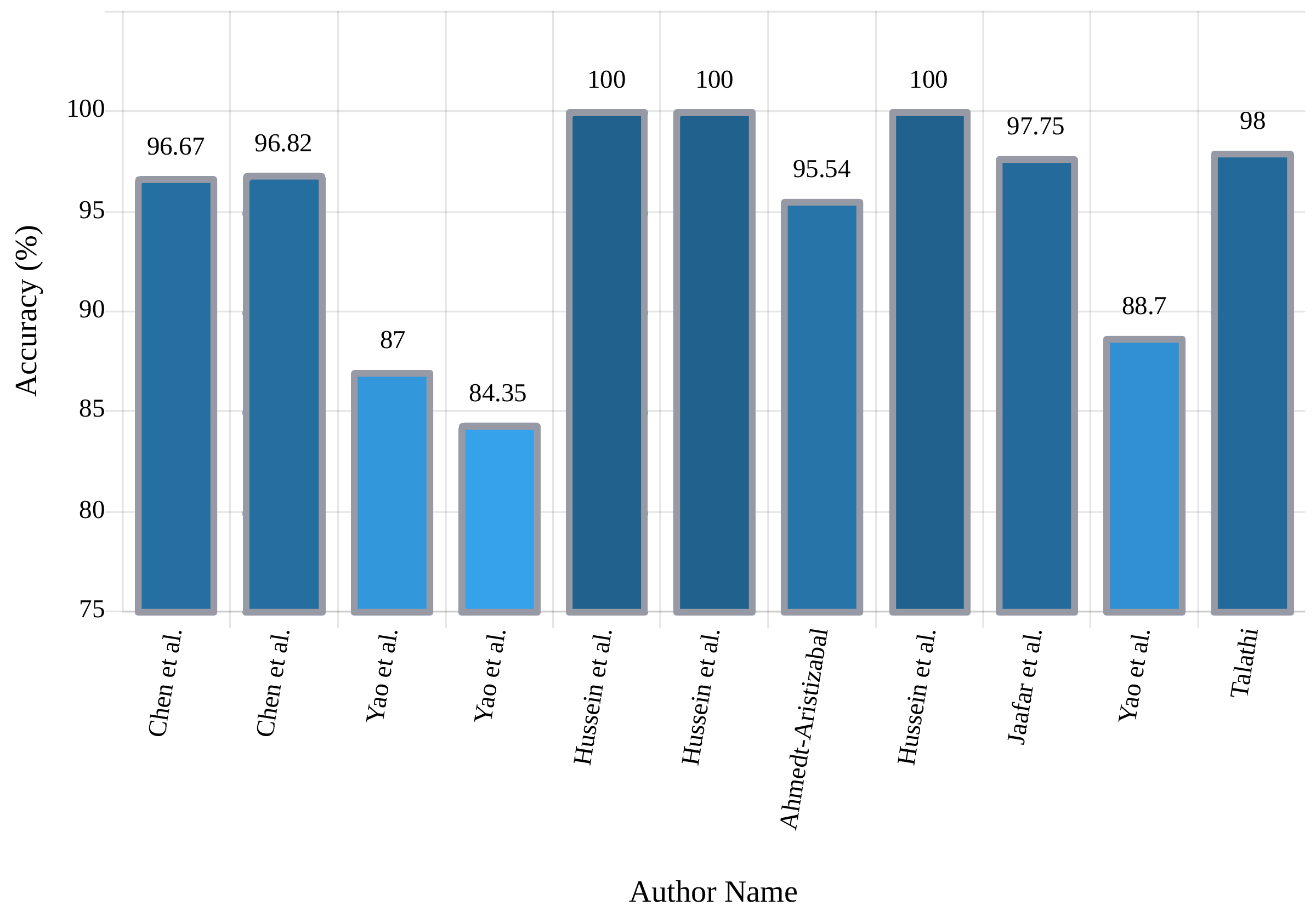}
    
    \caption{Sketch of accuracy (\%) versus authors obtained using RNN models for seizure detection.}
    \label{fig:rnn}
\end{figure}

One variation of LSTM is GRU, which combines the input and forgets gates into one update gate \cite{seventeen,eighteen,ninteen,twenty}.  It merges the input and forgets gates and also makes some other modifications. The gating signals are decreased to two. One is the reset gate, and another is the updating gate. These two gates decide which information is necessary to pass to the output. In one experiment, Chen et al. \cite{fiftyOne} used a 3-layer GRU network with sigmoid classification and yielded 96.67\% accuracy. A new GRU-based epileptic seizure detection system has been conducted by Talathi et al. \cite{eightyOne}. In the proposed technique, in the pre-processing step, the input signals were split into time windows and calculated from each spectrogram window and then applied to a 4-layer GRU network with a softmax FC layer in the classification stage and achieved 98\% accuracy. In another study, Roy et al. \cite{eightyTwo} employed a 5-layer GRU network and softmax classifier, achieved remarkable results. Table \ref{tablernn} provides the summary of related works done using RNNs. Figure \ref{fig:rnn} shows the sketch of accuracy (\%) obtained by various authors using RNN models for seizure detection.

\begin{table}[t]
\caption{Summary of related works done using RNNs.}
\label{tablernn}
\centering
\resizebox{3.5in}{!}{
\begin{tabular}{|c|c|c|c|c|}
\hline
\multicolumn{1}{|c|}{\multirow{2}{*}{Work}} & \multicolumn{1}{c|}{\multirow{2}{*}{Networks}} & \multicolumn{1}{c|}{\multirow{2}{*}{\shortstack[1]{Number \\of Layers}}} & \multicolumn{1}{c|}{\multirow{2}{*}{Classifier}} & \multicolumn{1}{c|}{\multirow{2}{*}{Accuracy\%}}\\
\multicolumn{1}{|c|}{} & \multicolumn{1}{c|}{} & \multicolumn{1}{c|}{} & \multicolumn{1}{c|}{} &  \multicolumn{1}{c|}{}\\
\hline
\multicolumn{1}{|c|}{\multirow{2}{*}{\shortstack[1]{\cite{sixtyFive}}}} & \multicolumn{1}{c|}{\multirow{2}{*}{\shortstack[1]{LSTM}}} & 3 & \multicolumn{1}{c|}{\multirow{2}{*}{\shortstack[1]{sigmoid}}} &  \multicolumn{1}{c|}{\multirow{2}{*}{\shortstack[1]{NA}}}\\
\cline{3-3}
\multicolumn{1}{|c|}{} & \multicolumn{1}{c|}{} & 4 & \multicolumn{1}{c|}{} &  \multicolumn{1}{c|}{}\\
\hline
% \multicolumn{1}{|c|}{\multirow{2}{*}{\shortstack[1]{\cite{seventySeven}}}} & \multicolumn{1}{c|}{\multirow{2}{*}{\shortstack[1]{GRU, LSTM,\\Simple RNN}}} & \multicolumn{1}{c|}{\multirow{2}{*}{\shortstack[1]{3}}} & \multicolumn{1}{c|}{\multirow{2}{*}{\shortstack[1]{LR}}} &  \multicolumn{1}{c|}{\multirow{2}{*}{\shortstack[1]{NA}}}\\
% \multicolumn{1}{|c|}{} & \multicolumn{1}{c|}{} & \multicolumn{1}{c|}{} & \multicolumn{1}{c|}{} &  \multicolumn{1}{c|}{}\\
% \hline
\cite{fiftyOne} & GRU & 3 & sigmoid & 96.67\\
\hline
\cite{fiftyOne} & LSTM & 3 & sigmoid & 96.82\\
\hline
\cite{fiftyFour} & 15-IndRNN & 48 & NA & 87.00\\
\hline
\cite{fiftyFour} & LSTM & 4 & NA & 84.35 \\
\hline
\multicolumn{1}{|c|}{\multirow{2}{*}{\shortstack[1]{\cite{fiftyNine}}}} & LSTM & \multicolumn{1}{c|}{\multirow{2}{*}{\shortstack[1]{6}}} & \multicolumn{1}{c|}{\multirow{2}{*}{\shortstack[1]{sigmoid}}} &  \multicolumn{1}{c|}{\multirow{2}{*}{\shortstack[1]{NA}}}\\
\cline{2-2}
\multicolumn{1}{|c|}{} & GRU & \multicolumn{1}{c|}{} & \multicolumn{1}{c|}{} &  \multicolumn{1}{c|}{}\\
\hline
\multicolumn{1}{|c|}{\multirow{3}{*}{\shortstack[1]{\cite{eightyThree}}}} & \multicolumn{1}{c|}{\multirow{3}{*}{\shortstack[1]{RNN}}} & \multicolumn{1}{c|}{\multirow{3}{*}{\shortstack[1]{NA}}} & \multicolumn{1}{c|}{\multirow{3}{*}{\shortstack[1]{MLP with 2 \\layers (logistic \\sigmoid Classifier)}}} &  \multicolumn{1}{c|}{\multirow{3}{*}{\shortstack[1]{NA}}}\\
\multicolumn{1}{|c|}{} & \multicolumn{1}{c|}{} & \multicolumn{1}{c|}{} & \multicolumn{1}{c|}{} &  \multicolumn{1}{c|}{}\\
\multicolumn{1}{|c|}{} & \multicolumn{1}{c|}{} & \multicolumn{1}{c|}{} & \multicolumn{1}{c|}{} &  \multicolumn{1}{c|}{}\\
\hline
\cite{eightyFour} & LSTM & 4 & softmax & 100 \\
\hline
\multicolumn{1}{|c|}{\multirow{4}{*}{\shortstack[1]{\cite{eightyFive}}}} & \multicolumn{1}{c|}{\multirow{4}{*}{\shortstack[1]{LSTM}}} & \multicolumn{1}{c|}{\multirow{2}{*}{\shortstack[1]{2}}} & \multicolumn{1}{c|}{\multirow{4}{*}{\shortstack[1]{sigmoid}}} &  \multicolumn{1}{c|}{\multirow{2}{*}{\shortstack[1]{95.54\\Validation}}}\\
\multicolumn{1}{|c|}{} & \multicolumn{1}{c|}{} & \multicolumn{1}{c|}{} & \multicolumn{1}{c|}{} &  \multicolumn{1}{c|}{}\\
\cline{3-3}
\cline{5-5}
\multicolumn{1}{|c|}{} & \multicolumn{1}{c|}{} & \multicolumn{1}{c|}{\multirow{2}{*}{\shortstack[1]{5}}} & \multicolumn{1}{c|}{} &  \multicolumn{1}{c|}{\multirow{2}{*}{\shortstack[1]{91.25\\Test}}}\\
\multicolumn{1}{|c|}{} & \multicolumn{1}{c|}{} & \multicolumn{1}{c|}{} & \multicolumn{1}{c|}{} &  \multicolumn{1}{c|}{}\\
\hline
\cite{eightySix} & LSTM & 4 & softmax & 100\\
\hline
% \cite{eightySeven} & LSTM & 4 & softmax & NA\\
% \hline
\cite{eightyEight} & LSTM & 3 & softmax & 97.75\\
\hline
\cite{eighty} & ADIndRNN-(3,3) & 31 & NA & 88.70 \\
\hline
\cite{eightyOne} & GRU & 4 & LR & 98.00\\
\hline
\cite{eightyTwo} & GRU & 5 & softmax & NA\\
\hline
\cite{hundredTwentyNine} & LSTM & 4 & softmax & 100\\
\hline

\end{tabular}}
\end{table}

\subsubsection{Autoencoders}

\textsc{Standard Autoencoders}

\begin{figure*}[ht]
    \centering
    \includegraphics[width=5.5in ]{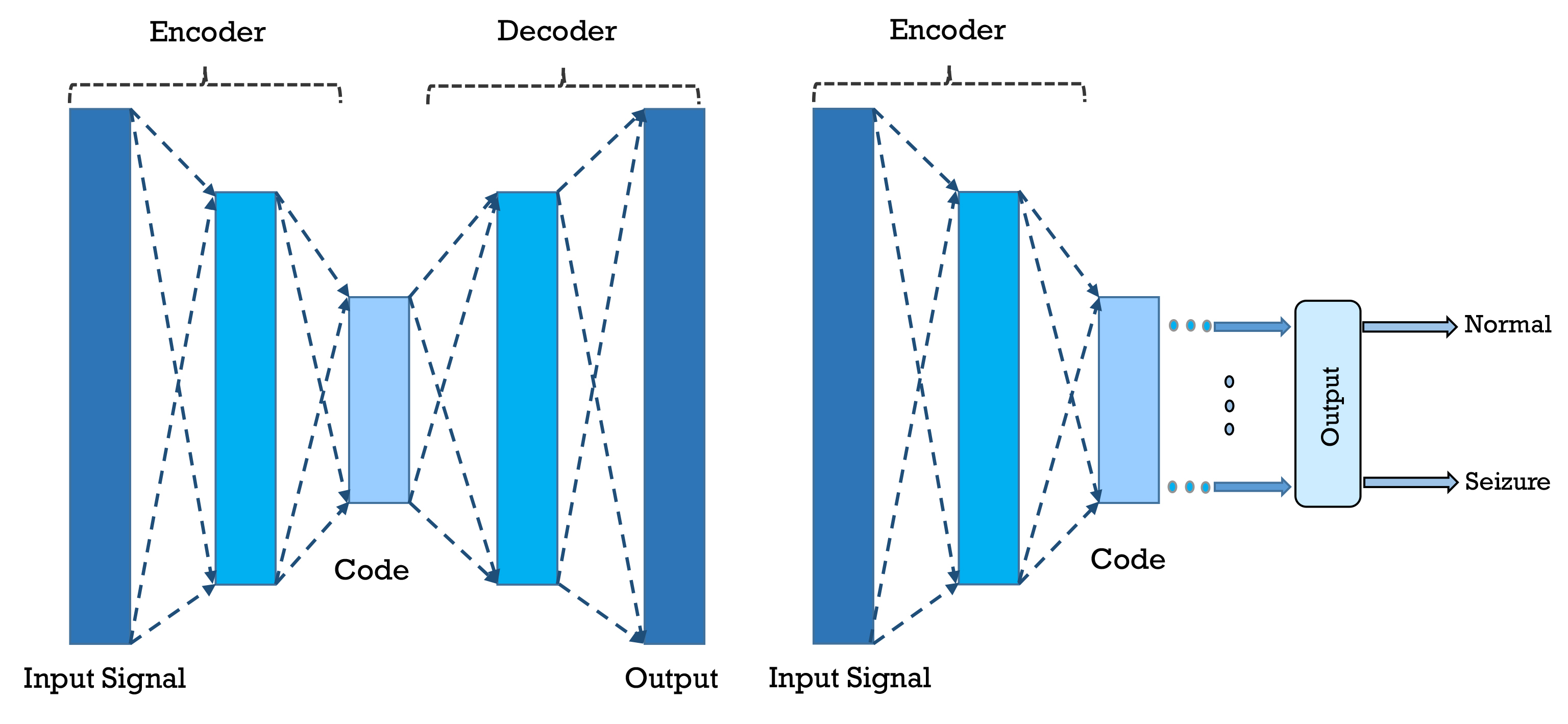}
    
    \caption{Sample AE network which may be used for seizure detection.}
    \label{fig:aefig}
\end{figure*}

AE is an unsupervised neural network machine learning model for which the input is the same as output \cite{seventeen,eighteen,ninteen,twenty}. Input is compressed to a latent-space representation, and then the output is obtained from the representation. So, in AE, the compression and decompression functions are coupled with the neural network. AE consists of three parts, i.e., encoder, code, and decoder. Autoencoder networks are the most commonly used for feature extraction or dimensionality reduction in brain signal processing. Figure \ref{fig:aefig} shows a general form of an AE used for epileptic seizure detection. As the first investigation in this section, Rajaguru et al. \cite{eightyNine} separately surveyed the Multilayer Autoencoders (MAE) and Expectation-Maximization with Principal Component Analysis (EM-PCA) methods to diminish the representation dimensions and then employed the Genetic Algorithm (GA) for classification. Experiments indicated that the average classification accuracy of 93.78\% is obtained when MAEs were applied for dimensionality reduction and combined with GA as classification. In another research, it was proposed to design an automated system based on AEs for the diagnosis of epilepsy using the EEG signal \cite{ninty}. First, Harmonic Wavelet Packet Transform (HWPT) was used to decompose the signal to frequency sub-bands, and then fractal features, including Box-Counting (BC), Multi-Resolution BC (MRBC) and Katz Fractal Dimension (KFD) were extracted from each of the sub-bands.

\textsc{Other Types of Autoencoders}

To create a more robust representation, a number of scheme have been applied to autoencoders \cite{goodfellow}, such as Denoising AE (DAE) (which tries to recreate input from a corrupted form of it) \cite{goodfellow}, Stacked AE (SAE) (stacking few autoencoders on top of each other to go deeper) \cite{goodfellow}, and Sparse AutoEncoders(SpAE) (which attempts to harness from sparse representations) \cite{goodfellow}. While these methods might pursue other objectives as well, for example, the DAE can be used to recover the corrupted input. However, the main goal in epileptic seizure detection systems is to prevent the hidden layer of autoencoder from merely learning the identity and finding more robust representation of the input. It is also common to combine two or more of these structures in an effort to get the best performance from them. 

Works in this section begin with Golmohammadi et al. \cite{sixtyFive}, who presented various deep networks, one of which is Stacked Denoising AE (SDAE). Their advised architecture in this section consists of three layers, and the final results demonstrate the proper performance of their approach. Qiu et al. \cite{nintyOne} exerted the windowed signal and z-score normalization step of pre-processing EEG signals and imported pre-processed data into the Denoising Sparse AE (DSpAE) network. In their experiment, they achieved an outstanding performance of 100\% accuracy. In \cite{nintyTwo}, a high-performance automated EEG analysis system based on principles of machine learning and big data is presented, which consists of several parts. At first, the signal features are extracted by Linear Predictive Cepstral Coefficients (LPCC) coefficients, then three paths are applied for precise detection. The first pass is sequential decoding using Hidden Markov Models (HMM), the second pass is composed of both temporal and spatial context analysis based on deep learning, in the third pass, a probabilistic grammar is employed.

\begin{figure}[t]
    \centering
    \includegraphics[width=3.5in]{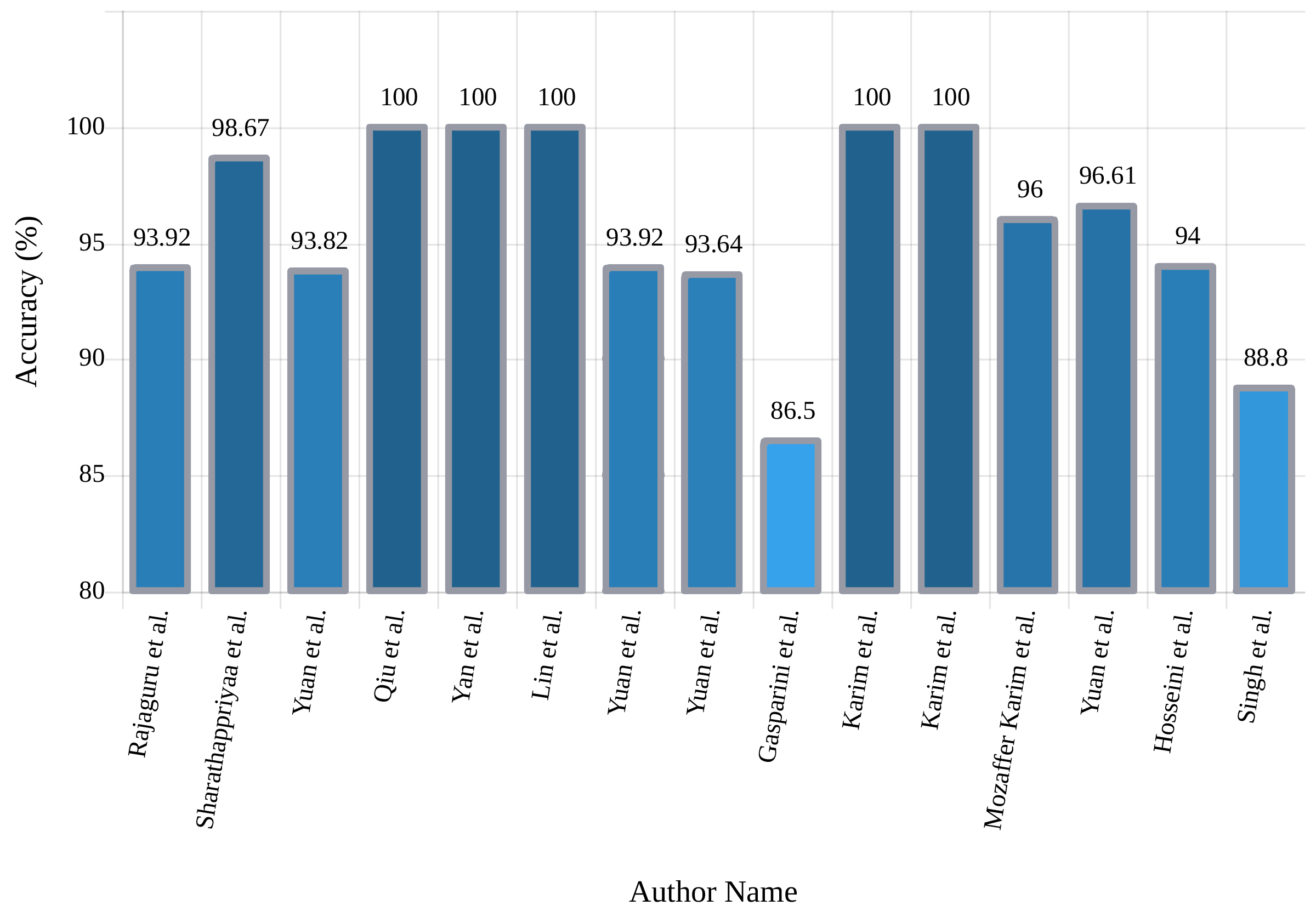}
    
    \caption{Sketch of accuracy (\%) versus authors obtained using AE models for seizure detection.}
    \label{fig:ae}
\end{figure}

In another research, Yan et al. \cite{nintyThree} proposed a feature extraction and classification method based on SpAE and Support Vector Machine (SVM). In this approach, first, the feature extraction of the input EEG signals is performed using SAE and, finally, the classification by SVM. Another SAE architecture was proposed by Yuan et al. \cite{nintyFour}, which is named Wave2Vec. In the pre-processing stage, the signals were first framed, and in the deep network segment, the SAE with softmax was applied and achieved 93.92\% accuracy. Following the experiments of Yuan et al., in \cite{nintyFive}, different Stacked Sparse Denoising AE (SSpDAE) architectures have been tested and compared. In this work, feature extraction is accomplished by the SSpDAE network and finally classification by softmax. They obtained an accuracy of 93.64\%. Table \ref{tableae} provides the summary of related works done using AEs. Also, Figure \ref{fig:ae} shows the comparison of the accuracies obtained by each of these investigations.

\begin{table}[t]
\caption{Summary of related works done using autoencoders.}
\label{tableae}
\centering
\resizebox{3.5in}{!}{
\begin{tabular}{|c|c|c|c|c|}
\hline
\multicolumn{1}{|c|}{\multirow{2}{*}{Work}} & \multicolumn{1}{c|}{\multirow{2}{*}{Networks}} & \multicolumn{1}{c|}{\multirow{2}{*}{\shortstack[1]{Number \\of Layers}}} & \multicolumn{1}{c|}{\multirow{2}{*}{Classifier}} & \multicolumn{1}{c|}{\multirow{2}{*}{Accuracy\%}}\\
\multicolumn{1}{|c|}{} & \multicolumn{1}{c|}{} & \multicolumn{1}{c|}{} & \multicolumn{1}{c|}{} &  \multicolumn{1}{c|}{}\\
\hline
\cite{sixtyFive} & SDAE & 3 & NA & NA\\
\hline
\cite{eightyNine} & MAE & NA & GA & 93.92\\
\hline
\cite{ninty} & AE & 3 & softmax & 98.67 \\
\hline
\cite{nintySix} & AE & One layer & sigmoid & NA\\
\hline
\multicolumn{1}{|c|}{\multirow{4}{*}{\shortstack[1]{\cite{nintySeven}}}} & \multicolumn{1}{c|}{\multirow{4}{*}{\shortstack[1]{SSpDAE}}} & \multicolumn{1}{c|}{\multirow{4}{*}{\shortstack[1]{2 hidden layers (intra\\channel) \& 3 hidden layer\\(cross channel) + 2 hidden\\layer (FC)+ classifier}}} & \multicolumn{1}{c|}{\multirow{4}{*}{\shortstack[1]{softmax}}} &  \multicolumn{1}{c|}{\multirow{4}{*}{\shortstack[1]{93.82}}}\\
\multicolumn{1}{|c|}{} & \multicolumn{1}{c|}{} & \multicolumn{1}{c|}{} & \multicolumn{1}{c|}{} &  \multicolumn{1}{c|}{}\\
\multicolumn{1}{|c|}{} & \multicolumn{1}{c|}{} & \multicolumn{1}{c|}{} & \multicolumn{1}{c|}{} &  \multicolumn{1}{c|}{}\\
\multicolumn{1}{|c|}{} & \multicolumn{1}{c|}{} & \multicolumn{1}{c|}{} & \multicolumn{1}{c|}{} &  \multicolumn{1}{c|}{}\\
\hline
\cite{nintyOne} & DSpAE & 3 & LR & 100\\
\hline
\multicolumn{1}{|c|}{\multirow{3}{*}{\shortstack[1]{\cite{nintyTwo}}}} & SPSW-SDA & \multicolumn{1}{c|}{\multirow{3}{*}{\shortstack[1]{Each model has\\3 hidden layers}}} & \multicolumn{1}{c|}{\multirow{3}{*}{\shortstack[1]{LR}}} &  \multicolumn{1}{c|}{\multirow{3}{*}{\shortstack[1]{NA}}}\\
\cline{2-2}
\multicolumn{1}{|c|}{} & 6W-SDA & \multicolumn{1}{c|}{} & \multicolumn{1}{c|}{} &  \multicolumn{1}{c|}{}\\
\cline{2-2}
\multicolumn{1}{|c|}{} & EYEM-SDA & \multicolumn{1}{c|}{} & \multicolumn{1}{c|}{} &  \multicolumn{1}{c|}{}\\
\hline
\cite{nintyThree} & SpAE & single-layer SpAE & SVM & 100\\
\hline
\cite{nintyEight} & SSpAE & 3-hidden-layer SSpAE & softmax & 100\\
\hline
\cite{nintyFour} & Wave2Vec & NA & softmax & 93.92 \\
\hline 
\cite{nintyFour} & SSpDAE & 2 & softmax & 93.64 \\
\hline
\cite{nintyNine} & SAE & 3 & softmax & 86.50\\
\hline
\cite{hundred} & SSpAE & 3 & softmax & 100 \\
\hline
\cite{hundredOne} & Deep SpAE & 3 & softmax & 100 \\
\hline
\cite{hundredTwo} & SAE & 3 (2 AE+ classifier) & softmax & 96.00\\
\hline
\cite{nintyFive} & SAE & 3 & softmax & 96.61 \\
\hline
\multicolumn{1}{|c|}{\multirow{2}{*}{\shortstack[1]{\cite{hundredThree}}}} & \multicolumn{1}{c|}{\multirow{2}{*}{\shortstack[1]{SSpAE}}} & \multicolumn{1}{c|}{\multirow{2}{*}{\shortstack[1]{3 (two sparse encoders as\\hidden layers+ classifier)}}} & \multicolumn{1}{c|}{\multirow{2}{*}{\shortstack[1]{softmax}}} &  \multicolumn{1}{c|}{\multirow{2}{*}{\shortstack[1]{94.00}}}\\
\multicolumn{1}{|c|}{} & \multicolumn{1}{c|}{} & \multicolumn{1}{c|}{} & \multicolumn{1}{c|}{} &  \multicolumn{1}{c|}{}\\
\hline
\cite{hundredFour} & SAE & 3 & softmax & 88.80 \\
\hline

\end{tabular}}
\end{table}

\subsubsection{Deep Belief and Boltzmann Networks}

Restricted Boltzmann Machines (RBM) is a variant of Deep Boltzmann Machines (DBM) and an undirected graphical model \cite{seventeen}. The unrestricted boltzmann machines may also have connections between the hidden units. Stacking the RBMs forms a DBN; RBM is the building block of DBN. DBNs are unsupervised probabilistic hybrid generative deep learning models comprising of latent and stochastic variables in multiple layers \cite{seventeen,eighteen}. Furthermore, a variation of DBN is called Convolutional DBN (CDBN), which could successfully scale the high dimensional model and uses the spatial information of the nearby pixels \cite{seventeen,eighteen}. Deep boltzmann machines are probabilistic, generative, unsupervised deep learning model which contains visible and multiple layers of hidden units \cite{seventeen,eighteen}. 

Xuyen et al. \cite{hundredFive} used DBN to identify epileptic spikes in EEG data. The proposed architecture in their study consisted of three hidden layers and achieved an accuracy of 96.87\%. In another study, Turner et al. \cite{hundredsix} applied the DBN network to diagnose epilepsy and found promising results. More information about DBN architecture for epileptic seizures is shown in table \ref{tabledbn}. 

\begin{table}[ht]
\caption{Summary of related works done using DBNs.}
\label{tabledbn}
\centering
\resizebox{3.5in}{!}{
\begin{tabular}{|c|c|c|c|c|}
\hline
\multicolumn{1}{|c|}{\multirow{2}{*}{Work}} & \multicolumn{1}{c|}{\multirow{2}{*}{Networks}} & \multicolumn{1}{c|}{\multirow{2}{*}{\shortstack[1]{Number \\of Layers}}} & \multicolumn{1}{c|}{\multirow{2}{*}{Classifier}} & \multicolumn{1}{c|}{\multirow{2}{*}{Accuracy\%}}\\
\multicolumn{1}{|c|}{} & \multicolumn{1}{c|}{} & \multicolumn{1}{c|}{} & \multicolumn{1}{c|}{} &  \multicolumn{1}{c|}{}\\
\hline
\cite{hundredFive} & DBN & 3 hidden layers & NA & 96.87\\
\hline
\cite{hundredsix} & DBN & 3 & LR & NA\\
\hline

\end{tabular}}
\end{table}

\subsubsection{CNN - RNN}

It is a highly efficient combination of deep learning networks to predict and diagnose epileptic seizures from EEG signals is the CNN-RNN architecture. Adding convolutional layers to RNN helps to find spatially nearby patterns more natural; meanwhile, RNN characteristic is more fitted for time-series processing. Starting with \cite{sixtyFive} as the first work in this section, they applied numerous pre-processing schemas; then, a modified CNN-LSTM architecture is proposed that consists of 13 layers and the sigmoid is used for the last layer. Finally, The proposed approach demonstrates high capability and performs competently. 

Roy et al. \cite{fourtySix} used different CNN-RNN hybrid architectures to improve the experimental results. Their first network comprised a one-dimensional 7-layer CNN-GRU convolution architecture, and the second one is a three-dimensional (3D) CNN-GRU network. In another work, Roy et al. \cite{eightyTwo} concentrated on natural and abnormal brain activities and suggested four different deep learning architectures. The proposed ChronoNet model was developed using previous models. It achieved 90.60\% and 86.57\% training and test accuracies respectively. 

\begin{figure}[t]
    \centering
    \includegraphics[width=3.5in]{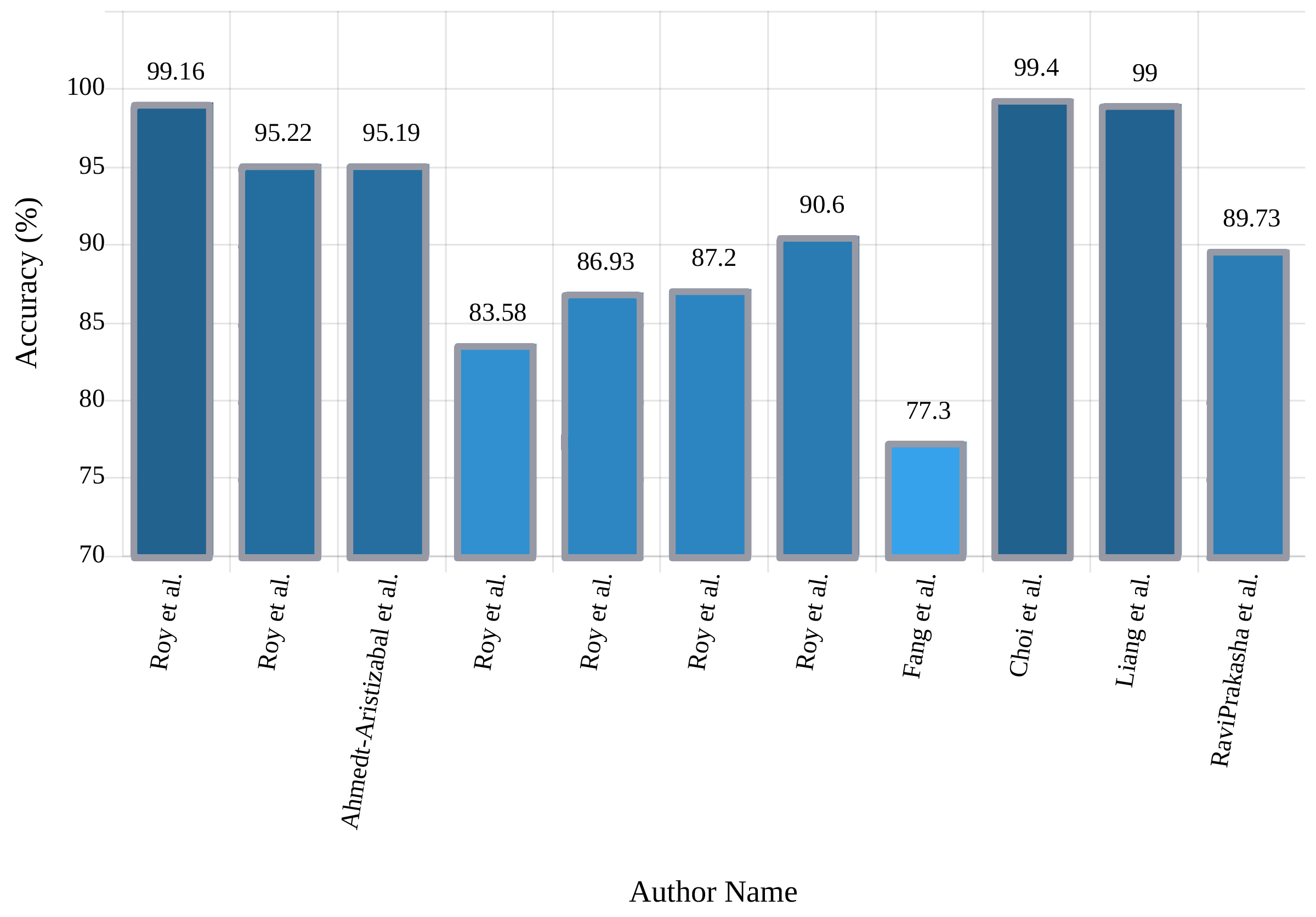}
    
    \caption{Sketch of accuracy (\%) versus authors obtained using CNN-RNN models for seizure detection.}
    \label{fig:cnnrnn}
\end{figure}

Fang et al. \cite{hundredSeven} used the Inception-V3 network. At the outset, a preliminary training was used on this network. Then, to fine-tune this architecture, an RNN based network called Spatial Temporal GRU (ST-GRU) CNN was applied, and they demonstrated that their approach achieved 77.30\% accuracy. Choi et al. \cite{hundredEight} proposed a Multi-scale 3D-CNN with an RNN model for the detection of epileptic seizures. The CNN module output is applied as the input of the RNN module. The RNN module consists of a unilateral GRU layer that extracts the temporal feature of epileptic seizures and is finally classified using an FC layer. At the end of this section, generalized information from the CNN-RNN research is presented in Table \ref{tablecnnrnn} and Figure \ref{fig:cnnrnn}, respectively.

\begin{table}[t]
\caption{Summary of related works done using CNN-RNNs.}
\label{tablecnnrnn}
\centering
\resizebox{3.5in}{!}{
\begin{tabular}{|c|c|c|c|c|}
\hline
\multicolumn{1}{|c|}{\multirow{2}{*}{Work}} & \multicolumn{1}{c|}{\multirow{2}{*}{Networks}} & \multicolumn{1}{c|}{\multirow{2}{*}{\shortstack[1]{Number \\of Layers}}} & \multicolumn{1}{c|}{\multirow{2}{*}{Classifier}} & \multicolumn{1}{c|}{\multirow{2}{*}{Accuracy\%}}\\
\multicolumn{1}{|c|}{} & \multicolumn{1}{c|}{} & \multicolumn{1}{c|}{} & \multicolumn{1}{c|}{} &  \multicolumn{1}{c|}{}\\
\hline
\multicolumn{1}{|c|}{\multirow{2}{*}{\shortstack[1]{\cite{sixtyFive}}}} & \multicolumn{1}{c|}{\multirow{2}{*}{\shortstack[1]{2D-CNN\\biLSTM}}} & \multicolumn{1}{c|}{\multirow{2}{*}{\shortstack[1]{13}}} & \multicolumn{1}{c|}{\multirow{2}{*}{\shortstack[1]{sigmoid}}} &  \multicolumn{1}{c|}{\multirow{2}{*}{\shortstack[1]{NA}}}\\
\multicolumn{1}{|c|}{} & \multicolumn{1}{c|}{} & \multicolumn{1}{c|}{} & \multicolumn{1}{c|}{} &  \multicolumn{1}{c|}{}\\
\hline
\cite{fourtySix} & 1D CNN-GRU & 7 & softmax & 99.16\\
\hline
\cite{fourtySix} & TCNN-RNN & 10 & softmax & 95.22\\
\hline
\cite{fourtyFour} & 2D CNN-LSTM & VGG-16 & sigmoid & 95.19\\
\hline
% \cite{seventyOne} & DCNN - Bi LSTM & 11 & sigmoid & 99.66\\
% \hline
\cite{eightyTwo} & C-RNN & 8 & softmax & 83.58\\
\hline
\cite{eightyTwo} & IC-RNN & 14 & softmax & 86.93 \\
\hline
\cite{eightyTwo} & C-DRNN & 8 & softmax & 87.20\\
\hline
\cite{eightyTwo} & ChronoNet & 14 & softmax & 90.60 \\
\hline
\cite{hundredNine} & 2D CNN-LSTM & 8 & NA & NA\\
\hline
\multicolumn{1}{|c|}{\multirow{3}{*}{\shortstack[1]{\cite{hundredSeven}}}} & \multicolumn{1}{c|}{\multirow{3}{*}{\shortstack[1]{ST-GRU ConvNets}}} & \multicolumn{1}{c|}{\multirow{3}{*}{\shortstack[1]{pre-trained\\Inception V3+\\GRU + FC}}} & \multicolumn{1}{c|}{\multirow{3}{*}{\shortstack[1]{NA}}} &  \multicolumn{1}{c|}{\multirow{3}{*}{\shortstack[1]{77.30}}}\\
\multicolumn{1}{|c|}{} & \multicolumn{1}{c|}{} & \multicolumn{1}{c|}{} & \multicolumn{1}{c|}{} &  \multicolumn{1}{c|}{}\\
\multicolumn{1}{|c|}{} & \multicolumn{1}{c|}{} & \multicolumn{1}{c|}{} & \multicolumn{1}{c|}{} &  \multicolumn{1}{c|}{}\\
\hline
\multicolumn{1}{|c|}{\multirow{2}{*}{\shortstack[1]{\cite{hundredEight}}}} & \multicolumn{1}{c|}{\multirow{2}{*}{\shortstack[1]{3D-CNN\\biGRU}}} & \multicolumn{1}{c|}{\multirow{2}{*}{\shortstack[1]{NA}}} & \multicolumn{1}{c|}{\multirow{2}{*}{\shortstack[1]{NA}}} &  \multicolumn{1}{c|}{\multirow{2}{*}{\shortstack[1]{99.40}}}\\
\multicolumn{1}{|c|}{} & \multicolumn{1}{c|}{} & \multicolumn{1}{c|}{} & \multicolumn{1}{c|}{} &  \multicolumn{1}{c|}{}\\
\hline
% \cite{hundredTen} & 2D CNN-LSTM & 17 & NA & NA\\
% \hline
% \multicolumn{1}{|c|}{\multirow{2}{*}{\shortstack[1]{\cite{hundredEleven}}}} & \multicolumn{1}{c|}{\multirow{2}{*}{\shortstack[1]{2D convolutional\\autoencoder-Bi LSTM}}} & \multicolumn{1}{c|}{\multirow{2}{*}{\shortstack[1]{29}}} & \multicolumn{1}{c|}{\multirow{2}{*}{\shortstack[1]{sigmoid}}} &  \multicolumn{1}{c|}{\multirow{2}{*}{\shortstack[1]{NA}}}\\
% \multicolumn{1}{|c|}{} & \multicolumn{1}{c|}{} & \multicolumn{1}{c|}{} & \multicolumn{1}{c|}{} &  \multicolumn{1}{c|}{}\\
% \hline
\cite{hundredTwelve} & 2D CNN-LSTM & 18 & softmax & 99.00\\
\hline
\multicolumn{1}{|c|}{\multirow{2}{*}{\shortstack[1]{\cite{hundredThirteen}}}} & \multicolumn{1}{c|}{\multirow{2}{*}{\shortstack[1]{1D CNN-LSTM}}} & 7 & \multicolumn{1}{c|}{\multirow{2}{*}{\shortstack[1]{sigmoid}}} &  \multicolumn{1}{c|}{\multirow{2}{*}{\shortstack[1]{89.73}}}\\
\cline{3-3}
\multicolumn{1}{|c|}{} & \multicolumn{1}{c|}{} & 8 & \multicolumn{1}{c|}{} &  \multicolumn{1}{c|}{}\\
\hline
\end{tabular}}
\end{table}

\subsubsection{CNN - AEs}

\begin{figure}[t]
    \centering
    \includegraphics[width=3.5in]{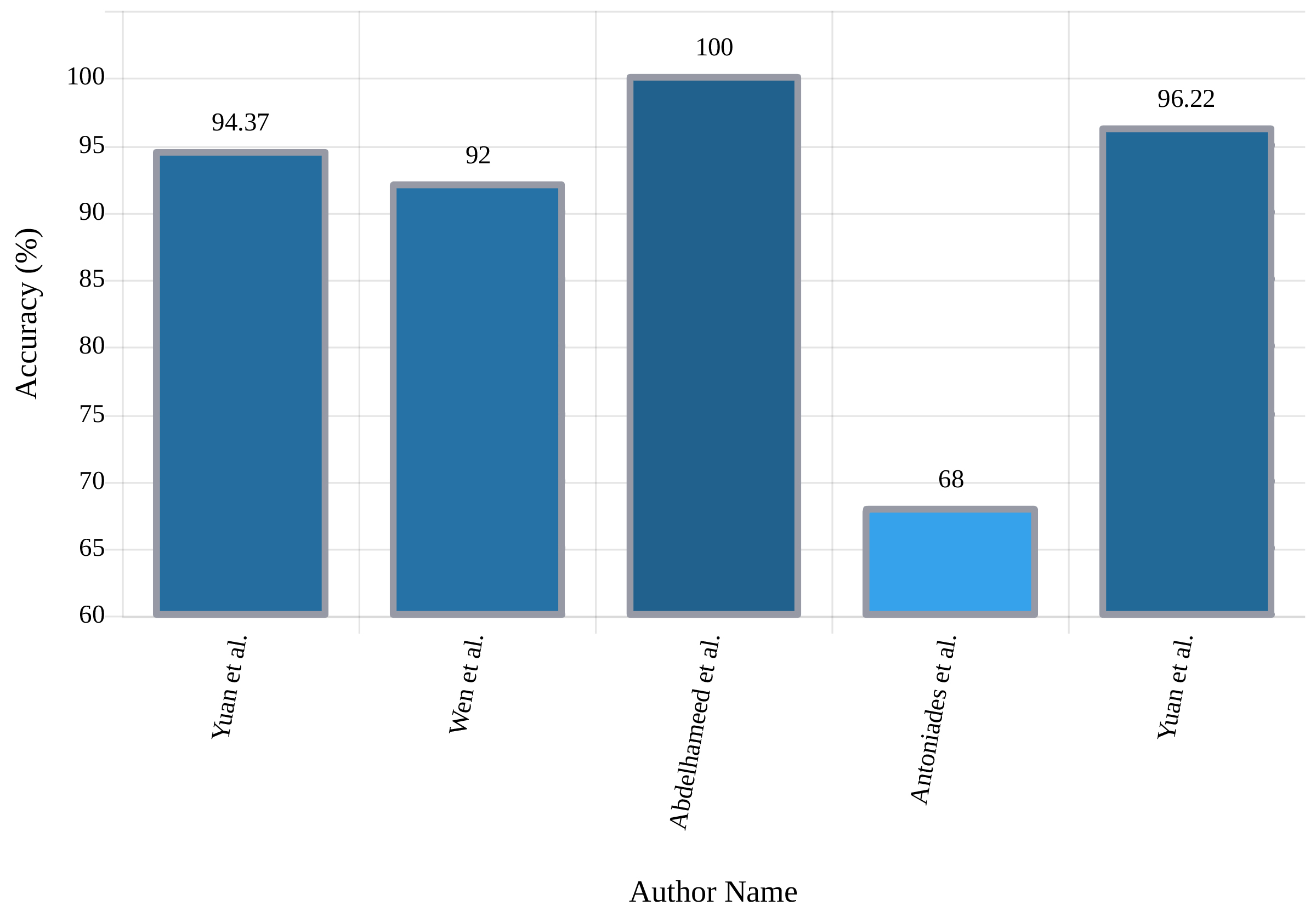}
    
    \caption{Sketch of accuracy (\%) versus authors obtained using CNN-AE models for seizure detection. }
    \label{fig:cnnae}
\end{figure}

In addition to finding nearby patterns, convolutional layers can reduce the number of parameters in structures such as autoencoders. These two reasons make their combination suitable for many tasks like unsupervised feature extraction for epileptic seizure detection. In this section, a novel approach based on CNN-AE was presented by Yuan et al. \cite{hundredFourteen}. At the feature extraction stage, two deep AE and 2D-CNN were used to extract the supervised and unsupervised features respectively. The unsupervised features were obtained directly from the input signals, and the supervised features were acquired from the spectrogram of the signals. Finally, the softmax classifier was utilized for classification and achieved 94.37\% accuracy. In another investigation, Yuan et al. \cite{hundredSixteen} proposed an approach called Deep Fusional Attention Network (DFAN) which can extract channel-aware representations from multi-channel EEG signals. They developed a fusional attention layer which utilized a fusional gate to fully integrate multi-view information to quantify the contribution of each biomedical channel dynamically. A multi-view convolution encoding layer, in combination with CNN, has also been used to train the integrated deep learning model. Table \ref{tablecnnae} provides the general information about the studies reviewed in the field of CNN-AE, and Figure \ref{fig:cnnae} shows the accuracy of each research.

\begin{table}[t]
\caption{Summary of related works done using CNN-AEs.}
\label{tablecnnae}
\centering
\resizebox{3.5in}{!}{
\begin{tabular}{|c|c|c|c|c|}
\hline
\multicolumn{1}{|c|}{\multirow{2}{*}{Work}} & \multicolumn{1}{c|}{\multirow{2}{*}{Networks}} & \multicolumn{1}{c|}{\multirow{2}{*}{\shortstack[1]{Number \\of Layers}}} & \multicolumn{1}{c|}{\multirow{2}{*}{Classifier}} & \multicolumn{1}{c|}{\multirow{2}{*}{Accuracy\%}}\\
\multicolumn{1}{|c|}{} & \multicolumn{1}{c|}{} & \multicolumn{1}{c|}{} & \multicolumn{1}{c|}{} &  \multicolumn{1}{c|}{}\\
\hline
% \cite{seventyOne} & DCAE + Bi LSTM & 19 & sigmoid & 99.66\\
% \hline
\cite{hundredFourteen} & CNN-AE & 10 & softmax & 94.37\\
\hline
\cite{hundredFifteen} & CNN-AE & 15 & Different Classifiers & 92.00\\
\hline
\multicolumn{1}{|c|}{\multirow{5}{*}{\shortstack[1]{\cite{hundredseventeen}}}} & \multicolumn{1}{c|}{\multirow{5}{*}{\shortstack[1]{1D CNN-AE\\(feature extraction)+\\MLP/LSTM/Bi\\LSTM (classification)}}} & \multicolumn{1}{c|}{\multirow{5}{*}{\shortstack[1]{16\\+\\3/1/1}}} & \multicolumn{1}{c|}{\multirow{2}{*}{\shortstack[1]{sigmoid}}} &  \multicolumn{1}{c|}{\multirow{2}{*}{\shortstack[1]{100\\2 classes}}}\\
\multicolumn{1}{|c|}{} & \multicolumn{1}{c|}{} & \multicolumn{1}{c|}{} & \multicolumn{1}{c|}{} &  \multicolumn{1}{c|}{}\\
\cline{4-5}
\multicolumn{1}{|c|}{} & \multicolumn{1}{c|}{} & \multicolumn{1}{c|}{} & \multicolumn{1}{c|}{\multirow{3}{*}{\shortstack[1]{softmax}}} &  \multicolumn{1}{c|}{\multirow{3}{*}{\shortstack[1]{99.33\\3 classes}}}\\
\multicolumn{1}{|c|}{} & \multicolumn{1}{c|}{} & \multicolumn{1}{c|}{} & \multicolumn{1}{c|}{} &  \multicolumn{1}{c|}{}\\
\multicolumn{1}{|c|}{} & \multicolumn{1}{c|}{} & \multicolumn{1}{c|}{} & \multicolumn{1}{c|}{} &  \multicolumn{1}{c|}{}\\
\hline
\multicolumn{1}{|c|}{\multirow{2}{*}{\shortstack[1]{\cite{hundredEighteen}}}} & CNN-ASAE & 8 & \multicolumn{1}{c|}{\multirow{2}{*}{\shortstack[1]{LR}}} & 66.00\\
\cline{2-3}
\cline{5-5}
\multicolumn{1}{|c|}{} & CNN-AAE & 7 & \multicolumn{1}{c|}{} &  68.00\\
\hline
% \multicolumn{1}{|c|}{\multirow{2}{*}{\shortstack[1]{\cite{hundredSixteen}}}} & \multicolumn{1}{c|}{\multirow{2}{*}{\shortstack[1]{Convolutional\\Encoder}}} & \multicolumn{1}{c|}{\multirow{2}{*}{\shortstack[1]{NA}}} & \multicolumn{1}{c|}{\multirow{2}{*}{\shortstack[1]{softmax}}} &  \multicolumn{1}{c|}{\multirow{2}{*}{\shortstack[1]{96.22}}}\\
% \multicolumn{1}{|c|}{} & \multicolumn{1}{c|}{} & \multicolumn{1}{c|}{} & \multicolumn{1}{c|}{} &  \multicolumn{1}{c|}{}\\
% \hline
\cite{hundredSixteen} & CNN-AE & NA & softmax & 96.22\\
\hline
\end{tabular}}
\end{table}

\section{Non-EEG Based Epileptic Seizure Detection}
\subsection{Medical Imaging Methods}

Various deep learning models were developed to detect epileptic seizure using MRI, structural MRI (sMRI), functional MRI (fMRI), resting-state fMRI (rs-fMRI) and PET scans with or without EEG signals \cite{hundredNinteen,hundredTwenty,hundredTwentyOne,hundredTwentyTwo,hundredTwentyThree,hundredTwentyFour,hundredTwentyFive,hundredTwentySix}. These models outperformed the conventional models in terms of automatic detection and monitoring of the disease. However, due to the nature and difficulties of using imaging methods, these models are mostly practiced for localization of seizure and detection is not the main aim of these models.

The authors \cite{hundredNinteen} proposed automatic localization and detection of Focal Cortical Dysplasia (FCD) from the MRI scan using a CNN. The FCD detection rate is only 50\% despite the progress in the analytics of MRI scans. Gill et al. \cite{hundredTwenty} proposed a CNN based algorithm with feature learning capability to detect FCD automatically. The researchers \cite{hundredTwentyOne} designed DeepIED based on deep learning and EEG-fMRI scans for epilepsy patients, combining the general linear model with EEG-fMRI techniques to estimate the epileptogenic zone. Hosseini et al. \cite{hundredTwentyTwo} proposed an edge computing autonomic framework for evaluation, regulation, and monitoring of the epileptic brain. The epileptogenic network was estimated using rs-fMRI and EEG. Shiri et al. \cite{hundredTwentySix} presented a technique for direct attenuation correction of PET images applying emission data only via a CNN-AE. Nineteen radiomic features from 83 brain regions were evaluated for image quantification via Hammersmith atlas. At the end of this section, generalized information from the MRI researches for seizures epileptic detection is presented in Table \ref{tablemri}.

\begin{table}[t]
\caption{Summary of related works done using medical imaging methods and deep learning.}
\label{tablemri}
\centering
\resizebox{3.5in}{!}{
\begin{tabular}{|c|c|c|c|c|}
\hline
\multicolumn{1}{|c|}{\multirow{2}{*}{Work}} & \multicolumn{1}{c|}{\multirow{2}{*}{Networks}} & \multicolumn{1}{c|}{\multirow{2}{*}{\shortstack[1]{Number \\of Layers}}} & \multicolumn{1}{c|}{\multirow{2}{*}{Classifier}} & \multicolumn{1}{c|}{\multirow{2}{*}{Accuracy\%}}\\
\multicolumn{1}{|c|}{} & \multicolumn{1}{c|}{} & \multicolumn{1}{c|}{} & \multicolumn{1}{c|}{} &  \multicolumn{1}{c|}{}\\
\hline
\cite{hundredNinteen} & 2D-CNN & 30 & sigmoid & 82.50\\
\hline
\cite{hundredTwenty} & 2D-CNN & 11 & softmax & NA\\
\hline
\multicolumn{1}{|c|}{\multirow{2}{*}{\shortstack[1]{\cite{hundredTwentyOne}}}} & \multicolumn{1}{c|}{\multirow{2}{*}{\shortstack[1]{ResNet}}} & \multicolumn{1}{c|}{\multirow{2}{*}{\shortstack[1]{31}}} & softmax &  \multicolumn{1}{c|}{\multirow{2}{*}{\shortstack[1]{NA}}}\\
\cline{4-4}
\multicolumn{1}{|c|}{} & \multicolumn{1}{c|}{} & \multicolumn{1}{c|}{} & Triplet &  \multicolumn{1}{c|}{}\\
\hline
\cite{hundredTwentyTwo} & 2D-CNN & NA & SVM & NA\\
\hline
\multicolumn{1}{|c|}{\multirow{2}{*}{\shortstack[1]{\cite{hundredTwentyThree}}}} & 2D-CNN & \multicolumn{1}{c|}{\multirow{2}{*}{\shortstack[1]{11}}} & \multicolumn{1}{c|}{\multirow{2}{*}{\shortstack[1]{softmax}}} &  89.80\\
\cline{2-2}
\cline{5-5}
\multicolumn{1}{|c|}{} & 3D-CNN & \multicolumn{1}{c|}{} & \multicolumn{1}{c|}{} &  82.50\\
\hline
\cite{hundredTwentyFour} & 2D-CNN & NA & NA & NA\\
\hline
\multicolumn{1}{|c|}{\multirow{4}{*}{\shortstack[1]{\cite{hundredTwentyFive}}}} & ResNet & \multicolumn{1}{c|}{\multirow{4}{*}{\shortstack[1]{14}}} & \multicolumn{1}{c|}{\multirow{4}{*}{\shortstack[1]{sigmoid}}} &  \multicolumn{1}{c|}{\multirow{4}{*}{\shortstack[1]{98.22}}}\\
\cline{2-2}
\multicolumn{1}{|c|}{} & VGGNet & \multicolumn{1}{c|}{} & \multicolumn{1}{c|}{} &  \multicolumn{1}{c|}{}\\
\cline{2-2}
\multicolumn{1}{|c|}{} & Inception-V3 & \multicolumn{1}{c|}{} & \multicolumn{1}{c|}{} &  \multicolumn{1}{c|}{}\\
\cline{2-2}
\multicolumn{1}{|c|}{} & SVGG-C3D & \multicolumn{1}{c|}{} & \multicolumn{1}{c|}{} &  \multicolumn{1}{c|}{}\\
\hline
\multicolumn{1}{|c|}{\multirow{4}{*}{\shortstack[1]{\cite{hundredTwentySix}}}} & \multicolumn{1}{c|}{\multirow{4}{*}{\shortstack[1]{Deep Direct\\Attenuation\\Correction (Deep-DAC)}}} & \multicolumn{1}{c|}{\multirow{4}{*}{\shortstack[1]{44}}} & \multicolumn{1}{c|}{\multirow{4}{*}{\shortstack[1]{tanh}}} &  \multicolumn{1}{c|}{\multirow{4}{*}{\shortstack[1]{NA}}}\\
\multicolumn{1}{|c|}{} & \multicolumn{1}{c|}{} & \multicolumn{1}{c|}{} & \multicolumn{1}{c|}{} &  \multicolumn{1}{c|}{}\\
\multicolumn{1}{|c|}{} & \multicolumn{1}{c|}{} & \multicolumn{1}{c|}{} & \multicolumn{1}{c|}{} &  \multicolumn{1}{c|}{}\\
\multicolumn{1}{|c|}{} & \multicolumn{1}{c|}{} & \multicolumn{1}{c|}{} & \multicolumn{1}{c|}{} &  \multicolumn{1}{c|}{}\\
\hline
\end{tabular}}
\end{table}

\subsection{Other Detection Methods}

Ravi Prakash et al. \cite{hundredThirteen} introduced an algorithm based on deep learning for ECoG based Functional Mapping (ECoG-FM) for eloquent language cortex identification. However, the success rate of ECoG-FM is low as compared to Electro-cortical Stimulation Mapping (ESM). In another work, Rosas-Romero et al. \cite{hundredTwentySeven} have used fNIRS to detect epileptic seizure and obtained better performance than using conventional EEG signals.

% HARDWARE USED FOR THE EPILEPTIC SEIZURE DETECTION
\section{Hardware And Software Used For The Epileptic Seizure Detection}
% \section{Epileptic Seizures Detection Hardware's (BCIs and Applications)}

\begin{figure}[t]
    \centering
    \includegraphics[width=3.5in ]{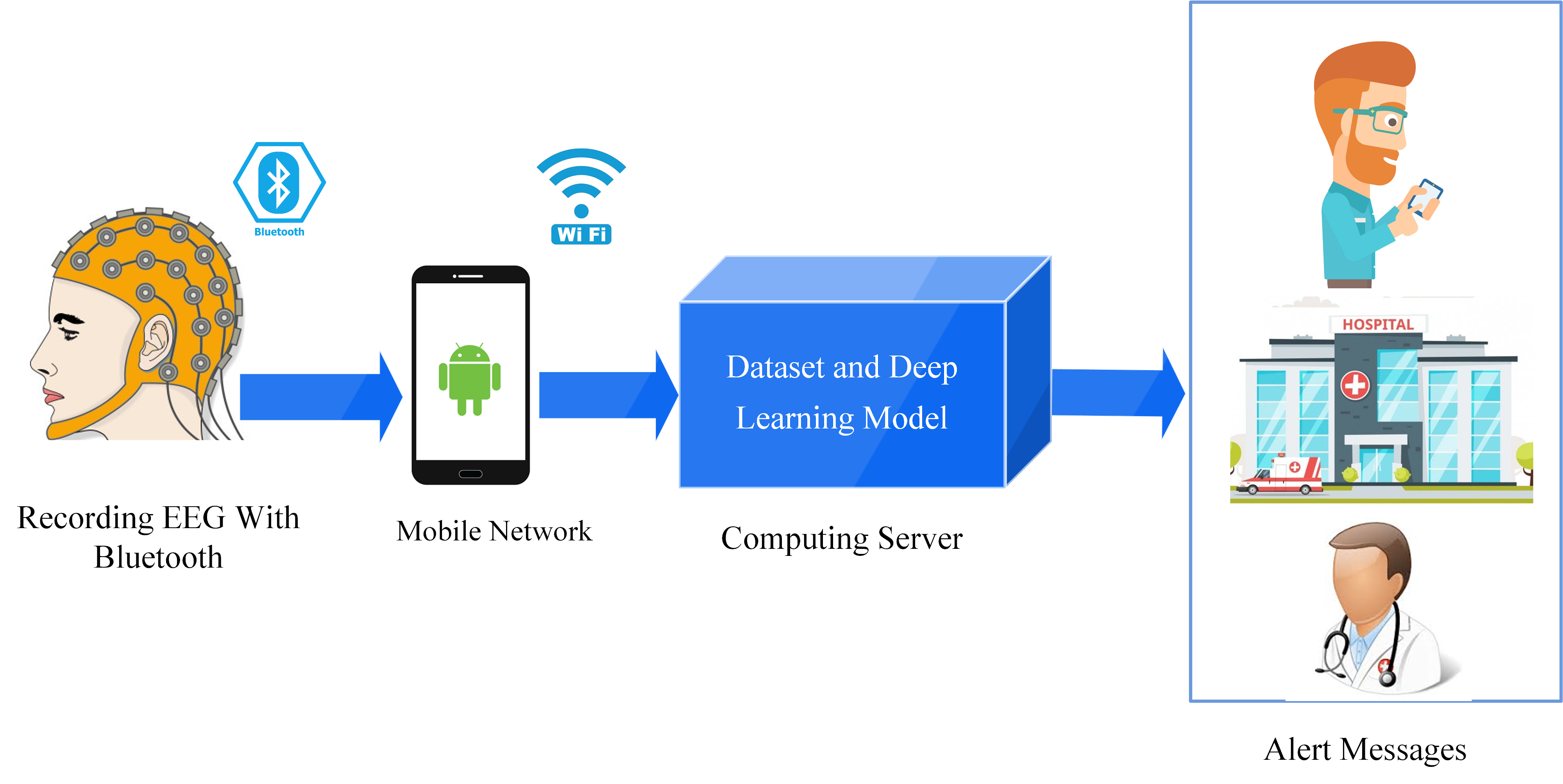}
    
    \caption{Block diagram of proposed epileptic seizure detection system using deep learning methods with EEG signals in \cite{hundredFour}.}
    \label{fig:tw}
\end{figure}

Fortunately, the privileged performance of deep learning algorithms has made them beneficial for commercial products. Nowadays various commercial products have been developed in the field of deep learning, one of which is deep learning applications and hardware for diagnosing epileptic seizures. In the first study investigated, the brain-computer interface (BCI) system was developed using an AE for epileptic seizure detection by Hosseini et al. \cite{hundredThree}. In another study, Singh et al. \cite{hundredFour} indicated a utilitarian product for the diagnosis of epileptic seizures, which comprised the user segment and the cloud segment. The block diagram of the proposed system presented by Singh et al. is shown in figure \ref{fig:tw}.

Kiral-Kornek et al. \cite{hundredTwentyEight} demonstrated that deep learning in combination with neuromorphic hardware could help in developing a wearable, real-time, always-on, patient-specific seizure warning system with low power consumption and reliable long-term performance. 

\section{Discussion}

Anticipating and timely recognition of epileptic seizures is of the essence, as it directly influences the quality of life of patients with this disease and can enhance their confidence in all life's stages. Numerous research has been fulfilled so far, concentrating on diagnosis epileptic seizures utilizing computer programs, but no efficient software programs or functional hardware have yet been implemented to recognize the disease; nevertheless, without an easy-to-use graphical interface, these studies might never be used in the real world. Until a few years ago, all the experimentation in this domain entailed traditional techniques, and due to practically the various deficiencies, they were not highly prosperous in assisting patients. As analyzed in this study, recent years of research into the diagnosis of epileptic seizures have led to the appearance of deep learning algorithms, and experts in the areas of artificial intelligence and signal processing reassure that these methods can sketch and lead to implementing concrete and functional tools. Table \ref{tablelast} in the Appendix yields more details of the research reviewed. It also shows what type of dataset, implementation tool, preprocessing, deep learning network, and evaluation method are utilized in the research. 

As shown in this study, various deep learning structures are applied for epileptic seizure detection, yet none of them has total superiority over others, and the best structure should be chosen carefully based on the dataset and problem characteristics, such as the need for real-time detection or minimum acceptable accuracy. As shown earlier, the obtained accuracies for detecting epileptic seizures are fair, and works in this field need to concentrate more on using clinical datasets and predicting epileptic seizures.
\section{Challenges}

Currently, challenges holding researchers back can be summarized as follows; firstly, many datasets only contain selected segments of EEG signals, which is not suitable for real-world applications where detection must be done from real-time signals, and clinical datasets are usually not available publicly. Secondly, while the number of datasets in this field is quite a lot, yet they can not be combined easily due to different frequency sampling or other parameters, making the size of total usable data for a model to train small. Lastly, deep learning models require massive computational resources, and these resources are not accessible to everyone. The researchers need to work in this direction to interfere with the epileptic seizures with better models, helping the patients at any time, anywhere.

\section{Conclusion and Future Works}

In this paper, a comprehensive review of researches in the field of epileptic seizure detection using numerous deep learning techniques such as CNNs, RNNs, and AEs is presented. Works studied in this research have used various screening methods, most importantly EEG and MRI. Finally, we have investigated deep learning based practical and applied hardware for diagnosing epileptic seizures. It is very encouraging that much of the future research will concentrate on hardware - practical applications to aid this kind of disease. The functional hardware has also been utilized to boost the performance of detection strategies. Furthermore, the models can be placed in the cloud by hospitals, so handheld applications, mobile or wearable devices, may be equipped with such models and the computations will be performed by cloud servers. The patients may also be benefited from predictive models for the epileptic seizure and take some measures to avert in a timely manner. Alert messages may be generated to the family, relatives, the concerned hospital, and doctor in the detection of epileptic seizures through the handheld devices or wearables, and thus the patient can be provided with proper treatment in time. Moreover, a cap with EEG electrodes in it can obtain the EEG signals and sent to the model kept in the cloud, so it can achieve real-time detection. Additionally, if we can detect interictal periods (early stage of seizure), early-stage immediately, the patient can take medication immediately and prevent seizure. Finally, this field of research requires more research by combining different screening methods for more precise and fast detection of epileptic seizures and also applying semi-supervised and unsupervised methods to further overcome the dataset size limits.

\appendices
\section{}
Table \ref{tablelast} shows details about all the works reviewed in this study.

\onecolumn
% \begin{table*}[t]
\begin{landscape}
\small
\begin{longtable}{|c|c|c|c|c|c|c|c|}
\caption{Summary of deep learning methods employed for automated detection of epileptic seizures.}    
    \label{tablelast}\\
%   \centering
%   \begin{tabular}{|c|c|c|c|c|c|c|c|}
  \hline
  Work & Dataset & Tools & Preprocessing & Network & K-fold & Classifier & Accuracy\%\\
  \hline
  \cite{twentyNine} & Clinical & NA & Spectrogram & 2D-CNN & NA & LR & 87.51\\
  \hline
%   \multicolumn{1}{|c|}{\multirow{3}{*}{\shortstack[1]{\cite{sixty}}}} & Freiburg & Python 2.7 & \multicolumn{1}{c|}{\multirow{3}{*}{\shortstack[1]{STFT}}} & 
%   \multicolumn{1}{c|}{\multirow{3}{*}{\shortstack[1]{2D-CNN}}} & 
%   \multicolumn{1}{c|}{\multirow{3}{*}{\shortstack[1]{NA}}} &
%   \multicolumn{1}{c|}{\multirow{3}{*}{\shortstack[1]{softmax}}} &\multicolumn{1}{c|}{\multirow{3}{*}{\shortstack[1]{NA}}}\\
%   \cline{2-3}
% \multicolumn{1}{|c|}{} & Kaggle & Keras & \multicolumn{1}{c|}{} & \multicolumn{1}{c|}{} & \multicolumn{1}{c|}{} & \multicolumn{1}{c|}{} & \multicolumn{1}{c|}{}\\
% \cline{2-3}
% \multicolumn{1}{|c|}{} & CHB-MIT & Tensorflow & \multicolumn{1}{c|}{} & \multicolumn{1}{c|}{} & \multicolumn{1}{c|}{} & \multicolumn{1}{c|}{} & \multicolumn{1}{c|}{}\\
% \hline
\cite{sixtyOne} & Clinical & MATLAB & Normalization & 2D-CNN & NA & softmax & NA\\
\hline
\multicolumn{1}{|c|}{\multirow{2}{*}{\shortstack[1]{\cite{sixtyTwo}}}} & Clinical & \multicolumn{1}{c|}{\multirow{2}{*}{\shortstack[1]{NA}}} & \multicolumn{1}{c|}{\multirow{2}{*}{\shortstack[1]{Filtering}}} & \multicolumn{1}{c|}{\multirow{2}{*}{\shortstack[1]{1D-CNN with\\2D-CNN}}} & \multicolumn{1}{c|}{\multirow{2}{*}{\shortstack[1]{NA}}} & \multicolumn{1}{c|}{\multirow{2}{*}{\shortstack[1]{sigmoid}}} & 90.50\\
\cline{2-2}
\cline{8-8}
\multicolumn{1}{|c|}{} & CHB-MIT & \multicolumn{1}{c|}{} & \multicolumn{1}{c|}{} & \multicolumn{1}{c|}{} & \multicolumn{1}{c|}{} & \multicolumn{1}{c|}{} & 85.60\\
\hline
% \multicolumn{1}{|c|}{\multirow{2}{*}{\shortstack[1]{\cite{sixtyThree}}}} & Clinical & \multicolumn{1}{c|}{\multirow{1}{*}{\shortstack[1]{Keras}}} & \multicolumn{1}{c|}{\multirow{2}{*}{\shortstack[1]{Filtering, BDS\\nonlinearity test, CWT}}} & \multicolumn{1}{c|}{\multirow{2}{*}{\shortstack[1]{2D-CNN}}} & \multicolumn{1}{c|}{\multirow{2}{*}{\shortstack[1]{NA}}} & \multicolumn{1}{c|}{\multirow{2}{*}{\shortstack[1]{softmax}}} & \multicolumn{1}{c|}{\multirow{2}{*}{\shortstack[1]{NA}}}\\
% \cline{2-3}
% \multicolumn{1}{|c|}{} & CHB-MIT & Theano & \multicolumn{1}{c|}{} & \multicolumn{1}{c|}{} & \multicolumn{1}{c|}{} & \multicolumn{1}{c|}{} & \multicolumn{1}{c|}{}\\
% \hline
\multicolumn{1}{|c|}{\multirow{3}{*}{\shortstack[1]{\cite{sixtyFour}}}} & \multicolumn{1}{c|}{\multirow{3}{*}{\shortstack[1]{Clinical}}} & Octave & \multicolumn{1}{c|}{\multirow{3}{*}{\shortstack[1]{Filtering, Re-referenced,\\Down Sampling}}} & \multicolumn{1}{c|}{\multirow{3}{*}{\shortstack[1]{2D-CNN}}} & \multicolumn{1}{c|}{\multirow{3}{*}{\shortstack[1]{NA}}} & \multicolumn{1}{c|}{\multirow{3}{*}{\shortstack[1]{softmax}}} & \multicolumn{1}{c|}{\multirow{3}{*}{\shortstack[1]{NA}}}\\
\cline{3-3}
\multicolumn{1}{|c|}{} & \multicolumn{1}{c|}{} & Keras & \multicolumn{1}{c|}{} & \multicolumn{1}{c|}{} & \multicolumn{1}{c|}{} & \multicolumn{1}{c|}{} & \multicolumn{1}{c|}{}\\
\cline{3-3}
\multicolumn{1}{|c|}{} & \multicolumn{1}{c|}{} & Theano & \multicolumn{1}{c|}{} & \multicolumn{1}{c|}{} & \multicolumn{1}{c|}{} & \multicolumn{1}{c|}{} & \multicolumn{1}{c|}{}\\
\hline
\multicolumn{1}{|c|}{\multirow{2}{*}{\shortstack[1]{\cite{sixtyFive}}}} & TUH EEG & \multicolumn{1}{c|}{\multirow{2}{*}{\shortstack[1]{NA}}} & \multicolumn{1}{c|}{\multirow{2}{*}{\shortstack[1]{Filtering}}} & \multicolumn{1}{c|}{\multirow{2}{*}{\shortstack[1]{CNN-RNN}}} & \multicolumn{1}{c|}{\multirow{2}{*}{\shortstack[1]{NA}}} & \multicolumn{1}{c|}{\multirow{2}{*}{\shortstack[1]{Different activation\\functions}}} & \multicolumn{1}{c|}{\multirow{2}{*}{\shortstack[1]{NA}}}\\
\cline{2-2}
\multicolumn{1}{|c|}{} & Clinical & \multicolumn{1}{c|}{} & \multicolumn{1}{c|}{} & \multicolumn{1}{c|}{} & \multicolumn{1}{c|}{} & \multicolumn{1}{c|}{} & \multicolumn{1}{c|}{}\\
\hline
\cite{fourtySix} & TUH EEG & NA & Different methods & 1D-CNN-GRU & NA & softmax & 99.16\\
\hline
\multicolumn{1}{|c|}{\multirow{2}{*}{\shortstack[1]{\cite{thirty}}}} & \multicolumn{1}{c|}{\multirow{2}{*}{\shortstack[1]{Clinical}}} & \multicolumn{1}{c|}{\multirow{2}{*}{\shortstack[1]{Keras}}} & \multicolumn{1}{c|}{\multirow{2}{*}{\shortstack[1]{Down sampling, Z-\\normalization, augmentation}}} & \multicolumn{1}{c|}{\multirow{2}{*}{\shortstack[1]{SeizNet}}} & \multicolumn{1}{c|}{\multirow{2}{*}{\shortstack[1]{NA}}} & \multicolumn{1}{c|}{\multirow{2}{*}{\shortstack[1]{NA}}} & \multicolumn{1}{c|}{\multirow{2}{*}{\shortstack[1]{NA}}}\\
\multicolumn{1}{|c|}{} & \multicolumn{1}{c|}{} & \multicolumn{1}{c|}{} & \multicolumn{1}{c|}{} & \multicolumn{1}{c|}{} & \multicolumn{1}{c|}{} & \multicolumn{1}{c|}{} & \multicolumn{1}{c|}{}\\
\hline
\multicolumn{1}{|c|}{\multirow{2}{*}{\shortstack[1]{\cite{thirtyOne}}}} & \multicolumn{1}{c|}{\multirow{2}{*}{\shortstack[1]{Clinical}}} & Python 3.6 & \multicolumn{1}{c|}{\multirow{2}{*}{\shortstack[1]{Z-Score normalization,\\STFT}}} & 1D-CNN & \multicolumn{1}{c|}{\multirow{2}{*}{\shortstack[1]{NA}}} & \multicolumn{1}{c|}{\multirow{2}{*}{\shortstack[1]{softmax}}} & \multicolumn{1}{c|}{\multirow{2}{*}{\shortstack[1]{NA}}}\\
\cline{3-3}
\cline{5-5}
\multicolumn{1}{|c|}{} & \multicolumn{1}{c|}{} & PyTorch & \multicolumn{1}{c|}{} & 2D-CNN & \multicolumn{1}{c|}{} & \multicolumn{1}{c|}{} & \multicolumn{1}{c|}{}\\
\hline
\cite{thirtyTwo} & CHB-MIT & PyTorch & Visualization & 2D-CNN & NA & softmax & 98.05\\
\hline
\multicolumn{1}{|c|}{\multirow{2}{*}{\shortstack[1]{\cite{thirtyThree}}}} & \multicolumn{1}{c|}{\multirow{2}{*}{\shortstack[1]{Clinical}}} & \multicolumn{1}{c|}{\multirow{2}{*}{\shortstack[1]{NA}}} & \multicolumn{1}{c|}{\multirow{2}{*}{\shortstack[1]{Filtering, Visualization,\\Normalization}}} & \multicolumn{1}{c|}{\multirow{2}{*}{\shortstack[1]{2D-CNN}}} & \multicolumn{1}{c|}{\multirow{2}{*}{\shortstack[1]{10}}} & \multicolumn{1}{c|}{\multirow{2}{*}{\shortstack[1]{softmax}}} & \multicolumn{1}{c|}{\multirow{2}{*}{\shortstack[1]{NA}}}\\
\multicolumn{1}{|c|}{} & \multicolumn{1}{c|}{} & \multicolumn{1}{c|}{} & \multicolumn{1}{c|}{} & \multicolumn{1}{c|}{} & \multicolumn{1}{c|}{} & \multicolumn{1}{c|}{} & \multicolumn{1}{c|}{}\\
\hline
\cite{thirtyFour} & TUH EEG & PyTorch & DivSpec & SeizureNet & 5 & softmax & NA\\
\hline
\multicolumn{1}{|c|}{\multirow{4}{*}{\shortstack[1]{\cite{fourtyFour}}}} & \multicolumn{1}{c|}{\multirow{4}{*}{\shortstack[1]{Clinical}}} & Caffe & \multicolumn{1}{c|}{\multirow{4}{*}{\shortstack[1]{Different Methods}}} & \multicolumn{1}{c|}{\multirow{2}{*}{\shortstack[1]{FRCNN with\\2D-CNN}}} & \multicolumn{1}{c|}{\multirow{4}{*}{\shortstack[1]{5}}} & \multicolumn{1}{c|}{\multirow{2}{*}{\shortstack[1]{SVM}}} & \multicolumn{1}{c|}{\multirow{4}{*}{\shortstack[1]{95.19}}}\\
\cline{3-3}
\multicolumn{1}{|c|}{} & \multicolumn{1}{c|}{} & OpenCV & \multicolumn{1}{c|}{} & \multicolumn{1}{c|}{} & \multicolumn{1}{c|}{} & \multicolumn{1}{c|}{} & \multicolumn{1}{c|}{}\\
\cline{3-3}
\cline{5-5}
\cline{7-7}
\multicolumn{1}{|c|}{} & \multicolumn{1}{c|}{} & \multicolumn{1}{c|}{\multirow{1}{*}{\shortstack[1]{Keras}}} & \multicolumn{1}{c|}{} & \multicolumn{1}{c|}{\multirow{2}{*}{\shortstack[1]{FRCNN with\\2D-CNN-LSTM}}} & \multicolumn{1}{c|}{} & \multicolumn{1}{c|}{\multirow{2}{*}{\shortstack[1]{sigmoid}}} & \multicolumn{1}{c|}{}\\
\cline{3-3}
\multicolumn{1}{|c|}{} & \multicolumn{1}{c|}{} & Theano & \multicolumn{1}{c|}{} & \multicolumn{1}{c|}{} & \multicolumn{1}{c|}{} & \multicolumn{1}{c|}{} & \multicolumn{1}{c|}{}\\
\hline
% \multicolumn{1}{|c|}{\multirow{2}{*}{\shortstack[1]{\cite{sixtySix}}}} & \multicolumn{1}{c|}{\multirow{2}{*}{\shortstack[1]{CHB-MIT}}} & \multicolumn{1}{c|}{\multirow{2}{*}{\shortstack[1]{NA}}} & \multicolumn{1}{c|}{\multirow{2}{*}{\shortstack[1]{DFT, MAS}}} & \multicolumn{1}{c|}{\multirow{2}{*}{\shortstack[1]{2D-CNN}}} & \multicolumn{1}{c|}{\multirow{2}{*}{\shortstack[1]{NA}}} & SVM & \multicolumn{1}{c|}{\multirow{2}{*}{\shortstack[1]{86.25}}}\\
% \cline{7-7}
% \multicolumn{1}{|c|}{} & \multicolumn{1}{c|}{} & \multicolumn{1}{c|}{} & \multicolumn{1}{c|}{} & \multicolumn{1}{c|}{} & \multicolumn{1}{c|}{} & softmax & \multicolumn{1}{c|}{}\\
% \hline
\cite{thirtyFive} & TUH EEG & TensorFlow & Feature Extraction & 2D-CNN & 10 & softmax & 74.00 \\
\hline
% \cite{sixtySeven} & Freiburg & MATLAB & LMD & 2D-CNN & NA & BLDA & 84.92\\
% \hline
\multicolumn{1}{|c|}{\multirow{2}{*}{\shortstack[1]{\cite{sixtyEight}}}} & Bern Barcelona & Octave & \multicolumn{1}{c|}{\multirow{2}{*}{\shortstack[1]{Filtering, EMD,\\DWT, Fourier}}} & \multicolumn{1}{c|}{\multirow{2}{*}{\shortstack[1]{2D-CNN}}} & \multicolumn{1}{c|}{\multirow{2}{*}{\shortstack[1]{5}}} & sigmoid & \multicolumn{1}{c|}{\multirow{2}{*}{\shortstack[1]{99.50}}}\\
\cline{2-2}
\cline{3-3}
\cline{7-7}
\multicolumn{1}{|c|}{} & Clinical & Keras & \multicolumn{1}{c|}{} & \multicolumn{1}{c|}{} & \multicolumn{1}{c|}{} & softmax & \multicolumn{1}{c|}{}\\
\hline
\cite{thirtySix} & Bern Barcelona & TensorFlow & STFT, Z-Score Normalization & 2D-CNN & 10 & softmax & 91.80\\
\hline
\cite{thirtySeven} & Clinical & NA & STFT & TGCN & NA & sigmoid & NA\\
\hline
\cite{thirtyEight} & Bonn & NA & DWT & 2D-CNN & 10 & softmax & 100\\
\hline
\cite{sixtyNine} & Bonn & Keras & CWT & 2D-CNN & 10 & softmax & 100\\
\hline
\multicolumn{1}{|c|}{\multirow{2}{*}{\shortstack[1]{\cite{thirtyNine}}}} & \multicolumn{1}{c|}{\multirow{2}{*}{\shortstack[1]{Bonn}}} & \multicolumn{1}{c|}{\multirow{2}{*}{\shortstack[1]{MATLAB}}} & \multicolumn{1}{c|}{\multirow{2}{*}{\shortstack[1]{Filtering}}} & \multicolumn{1}{c|}{\multirow{2}{*}{\shortstack[1]{2D-CNN}}} & \multicolumn{1}{c|}{\multirow{2}{*}{\shortstack[1]{NA}}} & \multicolumn{1}{c|}{\multirow{2}{*}{\shortstack[1]{softmax}}} & 99.60\\
\cline{8-8}
\multicolumn{1}{|c|}{} & \multicolumn{1}{c|}{} & \multicolumn{1}{c|}{} & \multicolumn{1}{c|}{} & \multicolumn{1}{c|}{} & \multicolumn{1}{c|}{} & \multicolumn{1}{c|}{} & 90.10\\
\hline
\multicolumn{1}{|c|}{\multirow{2}{*}{\shortstack[1]{\cite{seventy}}}} & \multicolumn{1}{c|}{\multirow{2}{*}{\shortstack[1]{CHB-MIT}}} & MATLAB & \multicolumn{1}{c|}{\multirow{2}{*}{\shortstack[1]{Over Sampling Method,\\FFT, WPD}}} & 2D-CNN & \multicolumn{1}{c|}{\multirow{2}{*}{\shortstack[1]{5}}} & \multicolumn{1}{c|}{\multirow{2}{*}{\shortstack[1]{MV-TSK-FS}}} & \multicolumn{1}{c|}{\multirow{2}{*}{\shortstack[1]{98.35}}}\\
\cline{3-3}
\cline{5-5}
\multicolumn{1}{|c|}{} & \multicolumn{1}{c|}{} & TensorFlow & \multicolumn{1}{c|}{} & 3D-CNN & \multicolumn{1}{c|}{} & \multicolumn{1}{c|}{} & \multicolumn{1}{c|}{}\\
\hline
\cite{fourty} & CHB-MIT & NA & Spatial Representation & 2D-CNN & NA & softmax & 99.48\\
\hline
% \multicolumn{1}{|c|}{\multirow{2}{*}{\shortstack[1]{\cite{seventyOne}}}} & \multicolumn{1}{c|}{\multirow{2}{*}{\shortstack[1]{CHB-MIT}}} & \multicolumn{1}{c|}{\multirow{2}{*}{\shortstack[1]{NA}}} & \multicolumn{1}{c|}{\multirow{2}{*}{\shortstack[1]{Visualization}}} & DCNN Bi-LSTM& \multicolumn{1}{c|}{\multirow{2}{*}{\shortstack[1]{NA}}} & \multicolumn{1}{c|}{\multirow{2}{*}{\shortstack[1]{sigmoid}}} & \multicolumn{1}{c|}{\multirow{2}{*}{\shortstack[1]{99.66}}}\\
% \cline{5-5}
% \multicolumn{1}{|c|}{} & \multicolumn{1}{c|}{} & \multicolumn{1}{c|}{} & \multicolumn{1}{c|}{} & DCAE Bi-LSTM & \multicolumn{1}{c|}{} & \multicolumn{1}{c|}{} & \multicolumn{1}{c|}{}\\
% \hline
\multicolumn{1}{|c|}{\multirow{2}{*}{\shortstack[1]{\cite{fourtyOne}}}} &
\multicolumn{1}{c|}{\multirow{2}{*}{\shortstack[1]{Clinical}}} & \multicolumn{1}{c|}{\multirow{2}{*}{\shortstack[1]{MATLAB}}} & \multicolumn{1}{c|}{\multirow{2}{*}{\shortstack[1]{Different Methods}}} & \multicolumn{1}{c|}{\multirow{2}{*}{\shortstack[1]{2D-CNN}}} & \multicolumn{1}{c|}{\multirow{2}{*}{\shortstack[1]{10}}} & sigmoid &  \multicolumn{1}{c|}{\multirow{2}{*}{\shortstack[1]{NA}}}\\
\cline{7-7}
\multicolumn{1}{|c|}{} & \multicolumn{1}{c|}{} & \multicolumn{1}{c|}{} & \multicolumn{1}{c|}{} & \multicolumn{1}{c|}{} & \multicolumn{1}{c|}{} & RF & \multicolumn{1}{c|}{}\\
\hline
\multicolumn{1}{|c|}{\multirow{2}{*}{\shortstack[1]{\cite{seventyTwo}}}} & CHB-MIT & \multicolumn{1}{c|}{\multirow{2}{*}{\shortstack[1]{NA}}} & \multicolumn{1}{c|}{\multirow{2}{*}{\shortstack[1]{MAS}}} & \multicolumn{1}{c|}{\multirow{2}{*}{\shortstack[1]{2D-CNN}}} & \multicolumn{1}{c|}{\multirow{2}{*}{\shortstack[1]{5}}} & \multicolumn{1}{c|}{\multirow{2}{*}{\shortstack[1]{KELM}}} & \multicolumn{1}{c|}{\multirow{2}{*}{\shortstack[1]{99.33}}}\\
\cline{2-2}
\multicolumn{1}{|c|}{} & Clinical & \multicolumn{1}{c|}{} & \multicolumn{1}{c|}{} & \multicolumn{1}{c|}{} & \multicolumn{1}{c|}{} & \multicolumn{1}{c|}{} & \multicolumn{1}{c|}{}\\
\hline
\cite{fourtyNine} & Clinical & TensorFlow & Filtering, Down Sampling & 1D-CNN & 4 & softmax, SVM & 83.86\\
\hline
\multicolumn{1}{|c|}{\multirow{3}{*}{\shortstack[1]{\cite{fourtyTwo}}}} & \multicolumn{1}{c|}{\multirow{3}{*}{\shortstack[1]{Bern Barcelona}}} & \multicolumn{1}{c|}{\multirow{3}{*}{\shortstack[1]{Caffe}}} & \multicolumn{1}{c|}{\multirow{3}{*}{\shortstack[1]{NA}}} & AlexNet & \multicolumn{1}{c|}{\multirow{3}{*}{\shortstack[1]{NA}}} & \multicolumn{1}{c|}{\multirow{3}{*}{\shortstack[1]{softmax}}} & \multicolumn{1}{c|}{\multirow{3}{*}{\shortstack[1]{100}}}\\
\cline{5-5}
\multicolumn{1}{|c|}{} & \multicolumn{1}{c|}{} & \multicolumn{1}{c|}{} & \multicolumn{1}{c|}{} & GoogleNet & \multicolumn{1}{c|}{} & \multicolumn{1}{c|}{} & \multicolumn{1}{c|}{}\\
\cline{5-5}
\multicolumn{1}{|c|}{} & \multicolumn{1}{c|}{} & \multicolumn{1}{c|}{} & \multicolumn{1}{c|}{} & LeNet & \multicolumn{1}{c|}{} & \multicolumn{1}{c|}{} & \multicolumn{1}{c|}{}\\
\hline
\cite{fourtyThree} & UCI & PyTorch & Signal2Image & 2D one Layer CNN & NA & DenseNet & 85.30\\
\hline
\cite{fourtyFive} & Clinical & Chainer & Filtering, Visualization & 2D-CNN & NA & softmax & NA\\
\hline
\cite{seventyThree} & Bonn & TensorFlow & Data Augmentation & P-1D-CNN & 10 & Majority Voting & 99.10\\
\hline
\cite{fourtySeven} & Bonn & MATLAB & Z-Score Normalization & 1D-CNN & 10 & softmax & 86.67\\
\hline
\cite{seventyFour} & CHB-MIT & NA & Filtering, Augmentation & MPCNN & NA & softmax & NA\\
\hline
\pagebreak
\hline
\cite{fourtyEight} & Clinical & Keras & Down-Sampling, Filtering & 1D-FCNN & 5 & softmax & NA\\
\hline
\cite{seventySix} & TUH EEG & Keras & Normalization and Standardization & 1D-CNN & NA & softmax & 79.34\\
\hline
\multicolumn{1}{|c|}{\multirow{2}{*}{\shortstack[1]{\cite{seventyFive}}}} & \multicolumn{1}{c|}{\multirow{2}{*}{\shortstack[1]{Clinical}}} & Theano & \multicolumn{1}{c|}{\multirow{2}{*}{\shortstack[1]{Filtering}}} & \multicolumn{1}{c|}{\multirow{2}{*}{\shortstack[1]{1D-CNN}}} & \multicolumn{1}{c|}{\multirow{2}{*}{\shortstack[1]{NA}}} & \multicolumn{1}{c|}{\multirow{2}{*}{\shortstack[1]{Binary LR}}} & \multicolumn{1}{c|}{\multirow{2}{*}{\shortstack[1]{NA}}}\\
\cline{3-3}
\multicolumn{1}{|c|}{} & \multicolumn{1}{c|}{} & Lasagne Library & \multicolumn{1}{c|}{} & \multicolumn{1}{c|}{} & \multicolumn{1}{c|}{} & \multicolumn{1}{c|}{} & \multicolumn{1}{c|}{}\\
\hline

% \multicolumn{1}{|c|}{\multirow{3}{*}{\shortstack[1]{\cite{seventySeven}}}} & \multicolumn{1}{c|}{\multirow{3}{*}{\shortstack[1]{Clinical}}} & Scipy & \multicolumn{1}{c|}{\multirow{3}{*}{\shortstack[1]{Different Methods}}} & 1D-CNN & \multicolumn{1}{c|}{\multirow{3}{*}{\shortstack[1]{NA}}} & \multicolumn{1}{c|}{\multirow{3}{*}{\shortstack[1]{LR}}} & \multicolumn{1}{c|}{\multirow{3}{*}{\shortstack[1]{NA}}}\\
% \cline{3-3}
% \cline{5-5}
% \multicolumn{1}{|c|}{} & \multicolumn{1}{c|}{} & Numpy & \multicolumn{1}{c|}{} & \multicolumn{1}{c|}{\multirow{2}{*}{\shortstack[1]{Different RNN\\Models}}} & \multicolumn{1}{c|}{} & \multicolumn{1}{c|}{} & \multicolumn{1}{c|}{}\\
% \cline{3-3}
% \multicolumn{1}{|c|}{} & \multicolumn{1}{c|}{} & Scikit-learn & \multicolumn{1}{c|}{} & \multicolumn{1}{c|}{} & \multicolumn{1}{c|}{} & \multicolumn{1}{c|}{} & \multicolumn{1}{c|}{}\\
% \hline
\multicolumn{1}{|c|}{\multirow{2}{*}{\shortstack[1]{\cite{fifty}}}} & \multicolumn{1}{c|}{\multirow{2}{*}{\shortstack[1]{CHB-MIT}}} & \multicolumn{1}{c|}{\multirow{2}{*}{\shortstack[1]{NA}}} & \multicolumn{1}{c|}{\multirow{2}{*}{\shortstack[1]{DWT, Feature Extraction,\\Normalization}}} & \multicolumn{1}{c|}{\multirow{2}{*}{\shortstack[1]{1D-CNN}}} & \multicolumn{1}{c|}{\multirow{2}{*}{\shortstack[1]{10}}} & \multicolumn{1}{c|}{\multirow{2}{*}{\shortstack[1]{NA}}} & \multicolumn{1}{c|}{\multirow{2}{*}{\shortstack[1]{99.07}}}\\
\multicolumn{1}{|c|}{} & \multicolumn{1}{c|}{} & \multicolumn{1}{c|}{} & \multicolumn{1}{c|}{} & \multicolumn{1}{c|}{} & \multicolumn{1}{c|}{} & \multicolumn{1}{c|}{} & \multicolumn{1}{c|}{}\\
\hline
\cite{fiftyOne} & Bonn & NA & DWT, Normalization & 1D-CNN & 5 & sigmoid & 97.27\\
\hline
\cite{fiftyTwo} & Bonn & NA & Normalization & 1D-TCNN & NA & NA & 100\\
\hline
\cite{fiftyThree} & Bonn & NA & EMD, MPF & 1D-CNN & 10 & softmax & 98.60\\
\hline
\cite{fiftyFour} & CHB-MIT & NA & Windowing & IndRNN & 10 & NA & 87.00\\
\hline
\multicolumn{1}{|c|}{\multirow{2}{*}{\shortstack[1]{\cite{seventyEight}}}} & Bonn & \multicolumn{1}{c|}{\multirow{2}{*}{\shortstack[1]{TensorFlow}}} & \multicolumn{1}{c|}{\multirow{2}{*}{\shortstack[1]{Filtering, Z-Score\\Normalization}}} & \multicolumn{1}{c|}{\multirow{2}{*}{\shortstack[1]{1D-CNN}}} & \multicolumn{1}{c|}{\multirow{2}{*}{\shortstack[1]{NA}}} & \multicolumn{1}{c|}{\multirow{2}{*}{\shortstack[1]{softmax}}} & 99.00\\
\cline{2-2}
\cline{8-8}
\multicolumn{1}{|c|}{} & Bern Barcelona & \multicolumn{1}{c|}{} & \multicolumn{1}{c|}{} & \multicolumn{1}{c|}{} & \multicolumn{1}{c|}{} & \multicolumn{1}{c|}{} & 91.80\\
\hline
\multicolumn{1}{|c|}{\multirow{2}{*}{\shortstack[1]{\cite{fiftyFive}}}} & CHB-MIT & \multicolumn{1}{c|}{\multirow{2}{*}{\shortstack[1]{PyTorch}}} & \multicolumn{1}{c|}{\multirow{2}{*}{\shortstack[1]{Filtering}}} & \multicolumn{1}{c|}{\multirow{2}{*}{\shortstack[1]{1D-PCM-CNN}}} & \multicolumn{1}{c|}{\multirow{2}{*}{\shortstack[1]{5}}} & \multicolumn{1}{c|}{\multirow{2}{*}{\shortstack[1]{softmax}}} & \multicolumn{1}{c|}{\multirow{2}{*}{\shortstack[1]{NA}}}\\
\cline{2-2}
\multicolumn{1}{|c|}{} & Clinical & \multicolumn{1}{c|}{} & \multicolumn{1}{c|}{} & \multicolumn{1}{c|}{} & \multicolumn{1}{c|}{} & \multicolumn{1}{c|}{} & \multicolumn{1}{c|}{}\\
\hline
\cite{seventyNine} & CHB-MIT & NA & MIDS, WGANs & 1D-CNN & NA & softmax & 84.00\\
\hline
\cite{fiftySix} & Clinical & NA & Down Sampling, PSD, FFT & 1D-CNN & 4 & sigmoid & 86.29\\
\hline
\cite{fiftySeven} & CHB-MIT & TensorFlow & Filtering & 1D-CNN & 4 & softmax & NA\\
\hline
\multicolumn{1}{|c|}{\multirow{3}{*}{\shortstack[1]{\cite{fiftyEight}}}} & \multicolumn{1}{c|}{\multirow{3}{*}{\shortstack[1]{NA}}} & Keras & \multicolumn{1}{c|}{\multirow{3}{*}{\shortstack[1]{Down Sampling, Filtering,\\Data Augmentation}}} & \multicolumn{1}{c|}{\multirow{3}{*}{\shortstack[1]{CNN-BP}}} & \multicolumn{1}{c|}{\multirow{3}{*}{\shortstack[1]{5}}} & \multicolumn{1}{c|}{\multirow{3}{*}{\shortstack[1]{sigmoid}}} & \multicolumn{1}{c|}{\multirow{3}{*}{\shortstack[1]{NA}}}\\
\cline{3-3}
\multicolumn{1}{|c|}{} & \multicolumn{1}{c|}{} & TensorFlow & \multicolumn{1}{c|}{} & \multicolumn{1}{c|}{} & \multicolumn{1}{c|}{} & \multicolumn{1}{c|}{} & \multicolumn{1}{c|}{}\\
\cline{3-3}
\multicolumn{1}{|c|}{} & \multicolumn{1}{c|}{} & MATLAB & \multicolumn{1}{c|}{} & \multicolumn{1}{c|}{} & \multicolumn{1}{c|}{} & \multicolumn{1}{c|}{} & \multicolumn{1}{c|}{}\\
\hline
\multicolumn{1}{|c|}{\multirow{3}{*}{\shortstack[1]{\cite{fiftyNine}}}} & \multicolumn{1}{c|}{\multirow{3}{*}{\shortstack[1]{Clinical}}} & \multicolumn{1}{c|}{\multirow{3}{*}{\shortstack[1]{NA}}} & \multicolumn{1}{c|}{\multirow{3}{*}{\shortstack[1]{Filtering, DWT}}} & 1D-CNN & \multicolumn{1}{c|}{\multirow{3}{*}{\shortstack[1]{NA}}} & sigmoid & \multicolumn{1}{c|}{\multirow{3}{*}{\shortstack[1]{NA}}}\\
\cline{5-5}
\cline{7-7}
\multicolumn{1}{|c|}{} & \multicolumn{1}{c|}{} & \multicolumn{1}{c|}{} & \multicolumn{1}{c|}{} & LSTM & \multicolumn{1}{c|}{} & RF & \multicolumn{1}{c|}{}\\
\cline{5-5}
\cline{7-7}
\multicolumn{1}{|c|}{} & \multicolumn{1}{c|}{} & \multicolumn{1}{c|}{} & \multicolumn{1}{c|}{} & GRU & \multicolumn{1}{c|}{} & SVM & \multicolumn{1}{c|}{}\\
\hline
\cite{eightyThree} & CHB-MIT & MATLAB & Filtering, Montage Mapping & DRNN & NA & MLP & NA\\
\hline
\cite{hundredTwentyNine} & Bonn & NA & Filtering & LSTM & NA & softmax & 100\\
\hline
\multicolumn{1}{|c|}{\multirow{3}{*}{\shortstack[1]{\cite{eightyFour}}}} & \multicolumn{1}{c|}{\multirow{3}{*}{\shortstack[1]{Bonn}}} & MATLAB & \multicolumn{1}{c|}{\multirow{3}{*}{\shortstack[1]{Filtering}}} & \multicolumn{1}{c|}{\multirow{3}{*}{\shortstack[1]{LSTM}}} & 3 & \multicolumn{1}{c|}{\multirow{3}{*}{\shortstack[1]{softmax}}} & \multicolumn{1}{c|}{\multirow{3}{*}{\shortstack[1]{100}}}\\
\cline{3-3}
\cline{6-6}
\multicolumn{1}{|c|}{} & \multicolumn{1}{c|}{} & Keras & \multicolumn{1}{c|}{} & \multicolumn{1}{c|}{} & 5 & \multicolumn{1}{c|}{} & \multicolumn{1}{c|}{}\\
\cline{3-3}
\cline{6-6}
\multicolumn{1}{|c|}{} & \multicolumn{1}{c|}{} & TensorFlow & \multicolumn{1}{c|}{} & \multicolumn{1}{c|}{} & 10 & \multicolumn{1}{c|}{} & \multicolumn{1}{c|}{}\\
\hline
\cite{eightyFive} & Bonn & Keras & Windowing & LSTM & 10 & sigmoid & 91.25\\
\hline
\multicolumn{1}{|c|}{\multirow{3}{*}{\shortstack[1]{\cite{eightySix}}}} & \multicolumn{1}{c|}{\multirow{3}{*}{\shortstack[1]{Bonn}}} & MATLAB & \multicolumn{1}{c|}{\multirow{3}{*}{\shortstack[1]{Filtering}}} & \multicolumn{1}{c|}{\multirow{3}{*}{\shortstack[1]{LSTM}}} & 3 & \multicolumn{1}{c|}{\multirow{3}{*}{\shortstack[1]{softmax}}} & \multicolumn{1}{c|}{\multirow{3}{*}{\shortstack[1]{100}}}\\
\cline{3-3}
\cline{6-6}
\multicolumn{1}{|c|}{} & \multicolumn{1}{c|}{} & Keras & \multicolumn{1}{c|}{} & \multicolumn{1}{c|}{} & 5 & \multicolumn{1}{c|}{} & \multicolumn{1}{c|}{}\\
\cline{3-3}
\cline{6-6}
\multicolumn{1}{|c|}{} & \multicolumn{1}{c|}{} & TensorFlow & \multicolumn{1}{c|}{} & \multicolumn{1}{c|}{} & 10 & \multicolumn{1}{c|}{} & \multicolumn{1}{c|}{}\\
\hline

% \multicolumn{1}{|c|}{\multirow{3}{*}{\shortstack[1]{\cite{eightySeven}}}} & \multicolumn{1}{c|}{\multirow{3}{*}{\shortstack[1]{CHB-MIT}}} & Python 3.6 & \multicolumn{1}{c|}{\multirow{3}{*}{\shortstack[1]{Feature Extraction}}} & \multicolumn{1}{c|}{\multirow{3}{*}{\shortstack[1]{LSTM}}} & 3 & \multicolumn{1}{c|}{\multirow{3}{*}{\shortstack[1]{softmax}}} & \multicolumn{1}{c|}{\multirow{3}{*}{\shortstack[1]{NA}}}\\
% \cline{3-3}
% \cline{6-6}
% \multicolumn{1}{|c|}{} & \multicolumn{1}{c|}{} & Keras & \multicolumn{1}{c|}{} & \multicolumn{1}{c|}{} & 5 & \multicolumn{1}{c|}{} & \multicolumn{1}{c|}{}\\
% \cline{3-3}
% \cline{6-6}
% \multicolumn{1}{|c|}{} & \multicolumn{1}{c|}{} & TensorFlow & \multicolumn{1}{c|}{} & \multicolumn{1}{c|}{} & 10 & \multicolumn{1}{c|}{} & \multicolumn{1}{c|}{}\\
% \hline
\multicolumn{1}{|c|}{\multirow{2}{*}{\shortstack[1]{\cite{eightyEight}}}} & \multicolumn{1}{c|}{\multirow{2}{*}{\shortstack[1]{Freiburg}}} & \multicolumn{1}{c|}{\multirow{2}{*}{\shortstack[1]{Anaconda\\Navigator}}} & \multicolumn{1}{c|}{\multirow{2}{*}{\shortstack[1]{Normalization, Filtering}}} & \multicolumn{1}{c|}{\multirow{2}{*}{\shortstack[1]{LSTM}}} & \multicolumn{1}{c|}{\multirow{2}{*}{\shortstack[1]{5}}} & \multicolumn{1}{c|}{\multirow{2}{*}{\shortstack[1]{softmax}}} & \multicolumn{1}{c|}{\multirow{2}{*}{\shortstack[1]{97.75}}}\\
\multicolumn{1}{|c|}{} & \multicolumn{1}{c|}{} & \multicolumn{1}{c|}{} & \multicolumn{1}{c|}{} & \multicolumn{1}{c|}{} & \multicolumn{1}{c|}{} & \multicolumn{1}{c|}{} & \multicolumn{1}{c|}{}\\
\hline
\multicolumn{1}{|c|}{\multirow{2}{*}{\shortstack[1]{\cite{eighty}}}} & CHB-MIT & \multicolumn{1}{c|}{\multirow{2}{*}{\shortstack[1]{NA}}} & \multicolumn{1}{c|}{\multirow{2}{*}{\shortstack[1]{Windowing}}} & \multicolumn{1}{c|}{\multirow{2}{*}{\shortstack[1]{ADIndRNN}}} & \multicolumn{1}{c|}{\multirow{2}{*}{\shortstack[1]{10}}} & \multicolumn{1}{c|}{\multirow{2}{*}{\shortstack[1]{NA}}} & \multicolumn{1}{c|}{\multirow{2}{*}{\shortstack[1]{88.70}}}\\
\cline{2-2}
\multicolumn{1}{|c|}{} & Bonn & \multicolumn{1}{c|}{} & \multicolumn{1}{c|}{} & \multicolumn{1}{c|}{} & \multicolumn{1}{c|}{} & \multicolumn{1}{c|}{} & \multicolumn{1}{c|}{}\\
\hline
\cite{eightyOne} & Bonn & Keras & Auto-Correlation & GRU & NA & LR & 98\\
\hline
\cite{eightyTwo} & TUH EEG & NA & TCP & ChronoNet & NA & softmax & 90.60\\
\hline
\cite{eightyNine} & Clinical & NA & Windowing & AE with EM-PCA & NA & GA & 93.92\\
\hline
\cite{ninty} & Bonn & MATLAB & Filtering, HWPT, FD & AE & NA & softmax & 98.67\\
\hline
\multicolumn{1}{|c|}{\multirow{2}{*}{\shortstack[1]{\cite{nintySix}}}} & \multicolumn{1}{c|}{\multirow{2}{*}{\shortstack[1]{Clinical}}} & \multicolumn{1}{c|}{\multirow{2}{*}{\shortstack[1]{TensorFlow}}} & \multicolumn{1}{c|}{\multirow{2}{*}{\shortstack[1]{Down Sampling, Filtering,\\Normalization}}} & \multicolumn{1}{c|}{\multirow{2}{*}{\shortstack[1]{AE}}} & \multicolumn{1}{c|}{\multirow{2}{*}{\shortstack[1]{NA}}} & \multicolumn{1}{c|}{\multirow{2}{*}{\shortstack[1]{sigmoid}}} & \multicolumn{1}{c|}{\multirow{2}{*}{\shortstack[1]{NA}}}\\
\multicolumn{1}{|c|}{} & \multicolumn{1}{c|}{} & \multicolumn{1}{c|}{} & \multicolumn{1}{c|}{} & \multicolumn{1}{c|}{} & \multicolumn{1}{c|}{} & \multicolumn{1}{c|}{} & \multicolumn{1}{c|}{}\\
\hline
\cite{nintySeven} & CHB-MIT & NA & STFT & SSDA & NA & softmax & 93.82\\
\hline
\multicolumn{1}{|c|}{\multirow{2}{*}{\shortstack[1]{\cite{nintyOne}}}} & \multicolumn{1}{c|}{\multirow{2}{*}{\shortstack[1]{Bonn}}} & \multicolumn{1}{c|}{\multirow{2}{*}{\shortstack[1]{MATLAB}}} & \multicolumn{1}{c|}{\multirow{2}{*}{\shortstack[1]{Z-Score Normalization,\\Standardization}}} & \multicolumn{1}{c|}{\multirow{2}{*}{\shortstack[1]{DSAE}}} & \multicolumn{1}{c|}{\multirow{2}{*}{\shortstack[1]{NA}}} & \multicolumn{1}{c|}{\multirow{2}{*}{\shortstack[1]{LR}}} & \multicolumn{1}{c|}{\multirow{2}{*}{\shortstack[1]{100}}}\\
\multicolumn{1}{|c|}{} & \multicolumn{1}{c|}{} & \multicolumn{1}{c|}{} & \multicolumn{1}{c|}{} & \multicolumn{1}{c|}{} & \multicolumn{1}{c|}{} & \multicolumn{1}{c|}{} & \multicolumn{1}{c|}{}\\
\hline
\multicolumn{1}{|c|}{\multirow{3}{*}{\shortstack[1]{\cite{nintyTwo}}}} & \multicolumn{1}{c|}{\multirow{3}{*}{\shortstack[1]{TUH EEG}}} & \multicolumn{1}{c|}{\multirow{2}{*}{\shortstack[1]{Open Source\\Toolkits}}} & \multicolumn{1}{c|}{\multirow{3}{*}{\shortstack[1]{Different Methods}}} & \multicolumn{1}{c|}{\multirow{3}{*}{\shortstack[1]{SDA}}} & \multicolumn{1}{c|}{\multirow{3}{*}{\shortstack[1]{NA}}} & \multicolumn{1}{c|}{\multirow{3}{*}{\shortstack[1]{LR}}} & \multicolumn{1}{c|}{\multirow{3}{*}{\shortstack[1]{NA}}}\\
\multicolumn{1}{|c|}{} & \multicolumn{1}{c|}{} & \multicolumn{1}{c|}{} & \multicolumn{1}{c|}{} & \multicolumn{1}{c|}{} & \multicolumn{1}{c|}{} & \multicolumn{1}{c|}{} & \multicolumn{1}{c|}{}\\
\cline{3-3}
\multicolumn{1}{|c|}{} & \multicolumn{1}{c|}{} & Theano & \multicolumn{1}{c|}{} & \multicolumn{1}{c|}{} & \multicolumn{1}{c|}{} & \multicolumn{1}{c|}{} & \multicolumn{1}{c|}{}\\
\hline
\cite{nintyThree} & Bonn & NA & Filtering & SAE & NA & SVM & 100\\
\hline
\pagebreak
\hline
\cite{nintyEight} & Bonn & NA & Normalization & SSAE & NA & softmax & 100\\
\hline
\cite{nintyFour} & CHB-MIT & Theano & Scalogram & Wave2Vec & NA & softmax & 93.92\\
\hline

\cite{hundredFourteen} & CHB-MIT & PyTorch & Data Augmentation, STFT & CNN-AE & 5 & softmax & 94.37\\
\hline
\multicolumn{1}{|c|}{\multirow{2}{*}{\shortstack[1]{\cite{nintyNine}}}} & \multicolumn{1}{c|}{\multirow{2}{*}{\shortstack[1]{Clinical}}} & \multicolumn{1}{c|}{\multirow{2}{*}{\shortstack[1]{NA}}} & \multicolumn{1}{c|}{\multirow{2}{*}{\shortstack[1]{Filtering, CWT,\\Feature Extraction}}} & \multicolumn{1}{c|}{\multirow{2}{*}{\shortstack[1]{SAE}}} & \multicolumn{1}{c|}{\multirow{2}{*}{\shortstack[1]{NA}}} & \multicolumn{1}{c|}{\multirow{2}{*}{\shortstack[1]{softmax}}} & \multicolumn{1}{c|}{\multirow{2}{*}{\shortstack[1]{86.50}}}\\
\multicolumn{1}{|c|}{} & \multicolumn{1}{c|}{} & \multicolumn{1}{c|}{} & \multicolumn{1}{c|}{} & \multicolumn{1}{c|}{} & \multicolumn{1}{c|}{} & \multicolumn{1}{c|}{} & \multicolumn{1}{c|}{}\\
\hline
\cite{hundred} & Bonn & NA & Taguchi Method & SSAE & NA & softmax & 100\\
\hline
\cite{hundredOne} & Clinical & NA & Dimension reduction, ESD & DeSAE & NA & softmax & 100\\
\hline
\cite{hundredTwo} & Bonn & NA & DWT & SAE & NA & softmax & 96.00\\
\hline
\cite{nintyFive} & CHB-MIT & NA & Different Methods & mSSDA & NA & softmax & 96.61\\
\hline
\cite{hundredThree} & Clinical & MATLAB & PCA, I-ICA & SSAE & NA & softmax & 94\\
\hline
\cite{hundredFour} & Bonn & MATLAB & Windowing & SAE & NA & softmax & 88.80\\
\hline
\cite{hundredFive} & Clinical & MATLAB & DWT & DBN & NA & NA & 96.87\\
\hline
\multicolumn{1}{|c|}{\multirow{3}{*}{\shortstack[1]{\cite{hundredsix}}}} & \multicolumn{1}{c|}{\multirow{3}{*}{\shortstack[1]{Clinical}}} & \multicolumn{1}{c|}{\multirow{3}{*}{\shortstack[1]{Theano}}} & \multicolumn{1}{c|}{\multirow{3}{*}{\shortstack[1]{Normalization, Feature\\Extraction, Standardization}}} & \multicolumn{1}{c|}{\multirow{3}{*}{\shortstack[1]{DBN}}} & \multicolumn{1}{c|}{\multirow{3}{*}{\shortstack[1]{NA}}} & LR & \multicolumn{1}{c|}{\multirow{3}{*}{\shortstack[1]{NA}}}\\
\cline{7-7}
\multicolumn{1}{|c|}{} & \multicolumn{1}{c|}{} & \multicolumn{1}{c|}{} & \multicolumn{1}{c|}{} & \multicolumn{1}{c|}{} & \multicolumn{1}{c|}{} & SVM & \multicolumn{1}{c|}{}\\
\cline{7-7}
\multicolumn{1}{|c|}{} & \multicolumn{1}{c|}{} & \multicolumn{1}{c|}{} & \multicolumn{1}{c|}{} & \multicolumn{1}{c|}{} & \multicolumn{1}{c|}{} & KNN & \multicolumn{1}{c|}{}\\
\hline
\cite{hundredNine} & CHB-MIT & NA & Image Based Representation & 2D CNN-LSTM & NA & NA & NA\\
\hline
\multicolumn{1}{|c|}{\multirow{2}{*}{\shortstack[1]{\cite{hundredSeven}}}} & \multicolumn{1}{c|}{\multirow{2}{*}{\shortstack[1]{Clinical}}} & \multicolumn{1}{c|}{\multirow{2}{*}{\shortstack[1]{TensorFlow}}} & \multicolumn{1}{c|}{\multirow{2}{*}{\shortstack[1]{Filtering}}} & \multicolumn{1}{c|}{\multirow{2}{*}{\shortstack[1]{ST-GRU\\ConvNets}}} & \multicolumn{1}{c|}{\multirow{2}{*}{\shortstack[1]{NA}}} & \multicolumn{1}{c|}{\multirow{2}{*}{\shortstack[1]{NA}}} & \multicolumn{1}{c|}{\multirow{2}{*}{\shortstack[1]{77.30}}}\\
\multicolumn{1}{|c|}{} & \multicolumn{1}{c|}{} & \multicolumn{1}{c|}{} & \multicolumn{1}{c|}{} & \multicolumn{1}{c|}{} & \multicolumn{1}{c|}{} & \multicolumn{1}{c|}{} & \multicolumn{1}{c|}{}\\
\hline
\multicolumn{1}{|c|}{\multirow{2}{*}{\shortstack[1]{\cite{hundredEight}}}} & CHB-MIT & \multicolumn{1}{c|}{\multirow{2}{*}{\shortstack[1]{NA}}} & \multicolumn{1}{c|}{\multirow{2}{*}{\shortstack[1]{STFT, 2D-mapping}}} & \multicolumn{1}{c|}{\multirow{2}{*}{\shortstack[1]{3D-CNN with\\Bi GRU}}} & \multicolumn{1}{c|}{\multirow{2}{*}{\shortstack[1]{NA}}} & \multicolumn{1}{c|}{\multirow{2}{*}{\shortstack[1]{NA}}} & \multicolumn{1}{c|}{\multirow{2}{*}{\shortstack[1]{99.40}}}\\
\cline{2-2}
\multicolumn{1}{|c|}{} & Clinical & \multicolumn{1}{c|}{} & \multicolumn{1}{c|}{} & \multicolumn{1}{c|}{} & \multicolumn{1}{c|}{} & \multicolumn{1}{c|}{} & \multicolumn{1}{c|}{}\\
\hline
% \multicolumn{1}{|c|}{\multirow{2}{*}{\shortstack[1]{\cite{hundredTen}}}} & \multicolumn{1}{c|}{\multirow{2}{*}{\shortstack[1]{CHB-MIT}}} & Keras & \multicolumn{1}{c|}{\multirow{2}{*}{\shortstack[1]{STFT, Standardization}}} & \multicolumn{1}{c|}{\multirow{2}{*}{\shortstack[1]{2D CNN-LSTM}}} & \multicolumn{1}{c|}{\multirow{2}{*}{\shortstack[1]{NA}}} & \multicolumn{1}{c|}{\multirow{2}{*}{\shortstack[1]{NA}}} & \multicolumn{1}{c|}{\multirow{2}{*}{\shortstack[1]{NA}}}\\
% \cline{3-3}
% \multicolumn{1}{|c|}{} & \multicolumn{1}{c|}{} & Theano & \multicolumn{1}{c|}{} & \multicolumn{1}{c|}{} & \multicolumn{1}{c|}{} & \multicolumn{1}{c|}{} & \multicolumn{1}{c|}{}\\
% \hline
% \multicolumn{1}{|c|}{\multirow{2}{*}{\shortstack[1]{\cite{hundredEleven}}}} & \multicolumn{1}{c|}{\multirow{2}{*}{\shortstack[1]{CHB-MIT}}} & \multicolumn{1}{c|}{\multirow{2}{*}{\shortstack[1]{NA}}} & \multicolumn{1}{c|}{\multirow{2}{*}{\shortstack[1]{Windowing}}} & \multicolumn{1}{c|}{\multirow{2}{*}{\shortstack[1]{2D-CNN with\\AE-Bi LSTM}}} & \multicolumn{1}{c|}{\multirow{2}{*}{\shortstack[1]{NA}}} & \multicolumn{1}{c|}{\multirow{2}{*}{\shortstack[1]{sigmoid}}} & \multicolumn{1}{c|}{\multirow{2}{*}{\shortstack[1]{NA}}}\\
% \multicolumn{1}{|c|}{} & \multicolumn{1}{c|}{} & \multicolumn{1}{c|}{} & \multicolumn{1}{c|}{} & \multicolumn{1}{c|}{} & \multicolumn{1}{c|}{} & \multicolumn{1}{c|}{} & \multicolumn{1}{c|}{}\\
% \hline
\cite{hundredTwelve} & CHB-MIT & NA & Visualization & 2D-CNN-LSTM & NA & softmax & 99.00\\
\hline
\cite{hundredThirteen} & clinical ECoG & NA & Filtering & 1D-CNN-LSTM & 5 & sigmoid & 89.73\\
\hline
\multicolumn{1}{|c|}{\multirow{2}{*}{\shortstack[1]{\cite{hundredFifteen}}}} & CHB-MIT & \multicolumn{1}{c|}{\multirow{2}{*}{\shortstack[1]{Scikit-Learn}}} & \multicolumn{1}{c|}{\multirow{2}{*}{\shortstack[1]{Channel Selection}}} & \multicolumn{1}{c|}{\multirow{2}{*}{\shortstack[1]{CNN-AE}}} & 5 & \multicolumn{1}{c|}{\multirow{2}{*}{\shortstack[1]{Different Methods}}} & \multicolumn{1}{c|}{\multirow{2}{*}{\shortstack[1]{92.00}}}\\
\cline{2-2}
\cline{6-6}
\multicolumn{1}{|c|}{} & Bonn & \multicolumn{1}{c|}{} & \multicolumn{1}{c|}{} & \multicolumn{1}{c|}{} & 10& \multicolumn{1}{c|}{} & \multicolumn{1}{c|}{}\\
\hline
\multicolumn{1}{|c|}{\multirow{2}{*}{\shortstack[1]{\cite{hundredseventeen}}}} & \multicolumn{1}{c|}{\multirow{2}{*}{\shortstack[1]{Bonn}}} & \multicolumn{1}{c|}{\multirow{2}{*}{\shortstack[1]{NA}}} & \multicolumn{1}{c|}{\multirow{2}{*}{\shortstack[1]{Windowing}}} & \multicolumn{1}{c|}{\multirow{2}{*}{\shortstack[1]{1D-CNN with\\Bi LSTM}}} & \multicolumn{1}{c|}{\multirow{2}{*}{\shortstack[1]{NA}}} & softmax & 99.33\\
\cline{7-8}
\multicolumn{1}{|c|}{} & \multicolumn{1}{c|}{} & \multicolumn{1}{c|}{} & \multicolumn{1}{c|}{} & \multicolumn{1}{c|}{} & \multicolumn{1}{c|}{} & sigmoid & 100\\
\hline
\multicolumn{1}{|c|}{\multirow{2}{*}{\shortstack[1]{\cite{hundredEighteen}}}} & \multicolumn{1}{c|}{\multirow{2}{*}{\shortstack[1]{Clinical}}} & \multicolumn{1}{c|}{\multirow{2}{*}{\shortstack[1]{Theano}}} & \multicolumn{1}{c|}{\multirow{2}{*}{\shortstack[1]{Mapping}}} & ASAE-CNN & \multicolumn{1}{c|}{\multirow{2}{*}{\shortstack[1]{NA}}} & \multicolumn{1}{c|}{\multirow{2}{*}{\shortstack[1]{LR}}} & 66.00\\
\cline{5-5}
\multicolumn{1}{|c|}{} & \multicolumn{1}{c|}{} & \multicolumn{1}{c|}{} & \multicolumn{1}{c|}{} & AAE-CNN & \multicolumn{1}{c|}{} & \multicolumn{1}{c|}{} & 68.00\\
\hline
\cite{hundredSixteen} & CHB-MIT & PyTorch & STFT & CNN-AE & 5 & softmax & 96.22\\
\hline
% \pagebreak
% \hline

\multicolumn{1}{|c|}{\multirow{3}{*}{\shortstack[1]{\cite{hundredNinteen}}}} & \multicolumn{1}{c|}{\multirow{3}{*}{\shortstack[1]{SCTIMST}}} & FSL & \multicolumn{1}{c|}{\multirow{3}{*}{\shortstack[1]{Noise reduction with BM3D\\algorithm, Skull-stripping,\\Segmentation, Postprocessing}}} & \multicolumn{1}{c|}{\multirow{3}{*}{\shortstack[1]{2D-CNN}}} &  \multicolumn{1}{c|}{\multirow{3}{*}{\shortstack[1]{5}}} & \multicolumn{1}{c|}{\multirow{3}{*}{\shortstack[1]{sigmoid}}} & \multicolumn{1}{c|}{\multirow{3}{*}{\shortstack[1]{NA}}}\\
\cline{3-3}
\multicolumn{1}{|c|}{} & \multicolumn{1}{c|}{} & Keras & \multicolumn{1}{c|}{} & \multicolumn{1}{c|}{} & \multicolumn{1}{c|}{} & \multicolumn{1}{c|}{} & \multicolumn{1}{c|}{}\\
\cline{3-3}
\multicolumn{1}{|c|}{} & \multicolumn{1}{c|}{} & TensorFlow & \multicolumn{1}{c|}{} & \multicolumn{1}{c|}{} & \multicolumn{1}{c|}{} & \multicolumn{1}{c|}{} & \multicolumn{1}{c|}{}\\
\hline
\cite{hundredTwenty} & Clinical MRI & NA & Different Methods & 2D-CNN & 5 & softmax & NA\\
\hline
\multicolumn{1}{|c|}{\multirow{2}{*}{\shortstack[1]{\cite{hundredTwentyOne}}}} & \multicolumn{1}{c|}{\multirow{2}{*}{\shortstack[1]{Clinical MRI}}} & \multicolumn{1}{c|}{\multirow{2}{*}{\shortstack[1]{Brain Vision\\Analyzer}}} & \multicolumn{1}{c|}{\multirow{2}{*}{\shortstack[1]{Filtering, ICA, BCG,\\GLM, MCS}}} & \multicolumn{1}{c|}{\multirow{2}{*}{\shortstack[1]{ResNet}}} & \multicolumn{1}{c|}{\multirow{2}{*}{\shortstack[1]{NA}}} & softmax & \multicolumn{1}{c|}{\multirow{2}{*}{\shortstack[1]{NA}}}\\
\cline{7-7}
\multicolumn{1}{|c|}{} & \multicolumn{1}{c|}{} & \multicolumn{1}{c|}{} & \multicolumn{1}{c|}{} & \multicolumn{1}{c|}{} & \multicolumn{1}{c|}{} & Triplet & \multicolumn{1}{c|}{}\\
\hline
\multicolumn{1}{|c|}{\multirow{3}{*}{\shortstack[1]{\cite{hundredTwentyTwo}}}} & ECoG Dataset & GIFT & \multicolumn{1}{c|}{\multirow{3}{*}{\shortstack[1]{Different Methods}}} & \multicolumn{1}{c|}{\multirow{3}{*}{\shortstack[1]{2D-CNN}}} & \multicolumn{1}{c|}{\multirow{3}{*}{\shortstack[1]{NA}}} & \multicolumn{1}{c|}{\multirow{3}{*}{\shortstack[1]{SVM}}} & \multicolumn{1}{c|}{\multirow{3}{*}{\shortstack[1]{NA}}}\\
\cline{2-3}
\multicolumn{1}{|c|}{} & \multicolumn{1}{c|}{\multirow{2}{*}{\shortstack[1]{Rs-FMRI\\Dataset}}} & FSL & \multicolumn{1}{c|}{} & \multicolumn{1}{c|}{} & \multicolumn{1}{c|}{} & \multicolumn{1}{c|}{} & \multicolumn{1}{c|}{}\\
\cline{3-3}
\multicolumn{1}{|c|}{} & \multicolumn{1}{c|}{} & FreeSurfer & \multicolumn{1}{c|}{} & \multicolumn{1}{c|}{} & \multicolumn{1}{c|}{} & \multicolumn{1}{c|}{} & \multicolumn{1}{c|}{}\\
\hline
\cite{hundredTwentyThree} & Clinical MRI & NA & Scaling Down & 3D-CNN & 5 & softmax & 89.80\\
\hline
\multicolumn{1}{|c|}{\multirow{2}{*}{\shortstack[1]{\cite{hundredTwentyFour}}}} & \multicolumn{1}{c|}{\multirow{2}{*}{\shortstack[1]{Clinical MRI}}} & \multicolumn{1}{c|}{\multirow{2}{*}{\shortstack[1]{FSL}}} & \multicolumn{1}{c|}{\multirow{2}{*}{\shortstack[1]{Connectivity Feature\\Extracion}}} & \multicolumn{1}{c|}{\multirow{2}{*}{\shortstack[1]{2D-CNN}}} & \multicolumn{1}{c|}{\multirow{2}{*}{\shortstack[1]{NA}}} & \multicolumn{1}{c|}{\multirow{2}{*}{\shortstack[1]{NA}}} & \multicolumn{1}{c|}{\multirow{2}{*}{\shortstack[1]{NA}}}\\
\multicolumn{1}{|c|}{} & \multicolumn{1}{c|}{} & \multicolumn{1}{c|}{} & \multicolumn{1}{c|}{} & \multicolumn{1}{c|}{} & \multicolumn{1}{c|}{} & \multicolumn{1}{c|}{} & \multicolumn{1}{c|}{}\\
\hline
\multicolumn{1}{|c|}{\multirow{4}{*}{\shortstack[1]{\cite{hundredTwentyFive}}}} & ImageNet & DPABI & \multicolumn{1}{c|}{\multirow{4}{*}{\shortstack[1]{ROI, Normalization,\\AAL, CNNI, Down-sampling,\\NNI (3D images)}}} & 2D-ResNet & \multicolumn{1}{c|}{\multirow{4}{*}{\shortstack[1]{NA}}} & \multicolumn{1}{c|}{\multirow{4}{*}{\shortstack[1]{sigmoid}}} & \multicolumn{1}{c|}{\multirow{4}{*}{\shortstack[1]{98.22}}}\\
\cline{2-3}
\cline{5-5}
\multicolumn{1}{|c|}{} & \multicolumn{1}{c|}{\multirow{2}{*}{\shortstack[1]{Pulmonary\\nodules Kaggle}}} & Python & \multicolumn{1}{c|}{} & 2D-VGGNET & \multicolumn{1}{c|}{} & \multicolumn{1}{c|}{} & \multicolumn{1}{c|}{}\\
\cline{3-3}
\cline{5-5}
\multicolumn{1}{|c|}{} & \multicolumn{1}{c|}{} & Keras & \multicolumn{1}{c|}{} & 2D-Inception V3 & \multicolumn{1}{c|}{} & \multicolumn{1}{c|}{} & \multicolumn{1}{c|}{}\\
\cline{2-3}
\cline{5-5}
\multicolumn{1}{|c|}{} & Clinical PET & TensorFlow & \multicolumn{1}{c|}{} & 3D-SVGG-C3D & \multicolumn{1}{c|}{} & \multicolumn{1}{c|}{} & \multicolumn{1}{c|}{}\\
\hline
\multicolumn{1}{|c|}{\multirow{2}{*}{\shortstack[1]{\cite{hundredTwentySix}}}} & \multicolumn{1}{c|}{\multirow{2}{*}{\shortstack[1]{Clinical PET}}} & \multicolumn{1}{c|}{\multirow{2}{*}{\shortstack[1]{TensorFlow}}} & \multicolumn{1}{c|}{\multirow{2}{*}{\shortstack[1]{OSEM, Data Augmentation\\Radionics Features}}} & \multicolumn{1}{c|}{\multirow{2}{*}{\shortstack[1]{DAC}}} & \multicolumn{1}{c|}{\multirow{2}{*}{\shortstack[1]{NA}}} & \multicolumn{1}{c|}{\multirow{2}{*}{\shortstack[1]{tanh}}} & \multicolumn{1}{c|}{\multirow{2}{*}{\shortstack[1]{NA}}}\\
\multicolumn{1}{|c|}{} & \multicolumn{1}{c|}{} & \multicolumn{1}{c|}{} & \multicolumn{1}{c|}{} & \multicolumn{1}{c|}{} & \multicolumn{1}{c|}{} & \multicolumn{1}{c|}{} & \multicolumn{1}{c|}{}\\
\hline
\end{longtable}

\end{landscape}
\twocolumn

% you can choose not to have a title for an appendix
% if you want by leaving the argument blank
% \section{}
% Appendix two text goes here.

% use section* for acknowledgment
% \section*{Acknowledgment}

% The authors would like to thank...

% Can use something like this to put references on a page
% by themselves when using endfloat and the captionsoff option.
% \ifCLASSOPTIONcaptionsoff
%   \newpage
% \fi

\bibliographystyle{IEEEtran}
\bibliography{main}

% % if you will not have a photo at all:
% \begin{IEEEbiographynophoto}{John Doe}
% Biography text here.
% \end{IEEEbiographynophoto}

% \begin{IEEEbiographynophoto}{Jane Doe}
% Biography text here.
% \end{IEEEbiographynophoto}

\end{document}